\documentclass[final]{cvpr}
\usepackage{times}
\usepackage{graphicx}
\usepackage{comment}
\usepackage{amsmath,amssymb} 
\usepackage{color}
\usepackage{epsfig}
\usepackage{graphicx}
\usepackage{amsmath}
\usepackage{amssymb}
\usepackage[T1]{fontenc}
\usepackage{multirow}
\usepackage{caption}
\usepackage{mathrsfs}
\usepackage{url}

\newcommand{\E}{\mathbb{E}}

\def\E{{\rm E}}

\usepackage[pagebackref=true,breaklinks=true,colorlinks,bookmarks=false]{hyperref}

\def\cvprPaperID{10581} 
\def\confYear{CVPR 2021}

\newcommand*\samethanks[1][\value{footnote}]{\footnotemark[#1]}

\begin{document}

\title{Generative PointNet: Deep Energy-Based Learning on Unordered Point Sets for 3D Generation, Reconstruction and Classification}

\author{Jianwen Xie $^{1}$\thanks{Equal contributions.} ,  Yifei Xu $^{2}$\samethanks[1] , Zilong Zheng $^2$, Song-Chun Zhu $^{2,3,4}$, Ying Nian Wu $^2$\\
$^1$ Cognitive Computing Lab, Baidu Research, Bellevue, WA, USA\\
$^2$ University of California, Los Angeles (UCLA), CA, USA\\
$^3$ Tsinghua University, Beijing, China\\
$^4$ Peking University, Beijing, China\\
{\tt\small \{jianwen, fei960922, z.zheng\}@ucla.edu,  \{sczhu, ywu\}@stat.ucla.edu}
}

\maketitle

\begin{abstract}
We propose a generative model of unordered point sets, such as point clouds, in the form of an energy-based model, where the energy function is parameterized by an input-permutation-invariant bottom-up neural network. The energy function learns a coordinate encoding of each point and then aggregates all individual point features into an energy for the whole point cloud. We call our model the Generative PointNet because it can be derived from the discriminative PointNet. Our model can be trained by MCMC-based maximum likelihood learning (as well as its variants), without the help of any assisting networks like those in GANs and VAEs. Unlike most point cloud generators that rely on hand-crafted distance metrics, our model does not require any hand-crafted distance metric for the point cloud generation, because it synthesizes point clouds by matching observed examples in terms of statistical properties defined by the energy function. Furthermore, we can learn a short-run MCMC toward the energy-based model as a flow-like generator for point cloud reconstruction and interpolation. The learned point cloud representation can be useful for point cloud classification. Experiments demonstrate the advantages of the proposed generative model of point clouds.  
\end{abstract}

\section{Introduction}
\subsection{Background and motivation}
Point clouds, as a standard 3D acquisition format used by devices like Lidar on autonomous vehicles, Kinect for Xbox and face identification sensor on phones, are getting increasingly popular for 3D representation in computer vision. Moreover, compared to other 3D formats such as voxel grids and 3D meshes, point clouds can provide a compact and detailed representation of a 3D object. 

Learning a generative model of 3D point clouds is a fundamental problem for 3D computer vision because it is beneficial to 3D point cloud synthesis and analysis tasks, by providing an explicit probability distribution of point clouds. Despite the enormous advance of discriminative models for the tasks of 3D point cloud classification and segmentation, e.g., PointNet~\cite{qi2017pointnet}, PointNet++~\cite{qi2017pointnet++}, DeepSet~\cite{zaheer2017deep},  ShapeContextNet~\cite{xie2018attentional}, PointGrid~\cite{le2018pointgrid}, DynamicGCN~\cite{wang2019dynamic}, and SampleNet~\cite{lang2020samplenet}, the progress in developing generative models for 3D point clouds has been lagging. A major challenge in generative modeling of point clouds is that unlike images, videos and volumetric shapes, point clouds are not regular structures but unordered point sets, which makes extending existing paradigms intended for structured data not straightforward. That is why the majority of existing works on 3D generative models are based on volumetric data, e.g., 3D ShapeNet~\cite{wu20153d}, 3D GAN~\cite{wu2016learning}, Generative VoxelNet~\cite{xie2018learning,Xie2020GenerativeVL}, 3D-INN~\cite{huang20193d}, etc.

With the recent success of a variety of generation tasks such as image generation and  video generation, researchers have become increasingly interested in point cloud generation, e.g., \cite{gadelha2018multiresolution,zamorski2018adversarial,valsesia2018learning,achlioptas2018learning,li2018point,yang2019pointflow}. Most of them are based on well-established frameworks of GAN \cite{goodfellow2014generative}  (e.g., \cite{zamorski2018adversarial,valsesia2018learning,achlioptas2018learning,li2018point}), VAE \cite{kingma2013auto} (e.g., \cite{gadelha2018multiresolution,yang2019pointflow}), or encoder-decoder with hand-crafted distance metrics, such as Chamfer distance or earth mover's distance \cite{fan2017point} for measuring the dissimilarity of two point clouds (e.g., \cite{gadelha2018multiresolution,zamorski2018adversarial}).   
In this paper, we propose a principled generative model for probabilistic modeling of 3D point clouds.  Specifically, the model is a probability density function directly defined on unordered point sets, and it is in the form of a deep energy-based model (EBM) \cite{xie2016theory} with the energy function parameterized by an input-permutation-invariant bottom-up deep network that is suitable for defining energy on an unordered point set. We call the proposed model the Generative PointNet because, following the theory presented in \cite{xie2016theory}, such a model can be derived from the discriminative PointNet \cite{qi2017pointnet}. The maximum likelihood estimation (MLE) of our model follows what Grenander \cite{grenander2007pattern} called ``analysis by synthesis'' scheme in pattern theory \cite{grenander1970unified}. Specifically, within each learning iteration, ``fake'' 3D point cloud examples are generated by Langevin dynamics sampling, which is a gradient-based Markov chain Monte Carlo \cite{liu2008monte,barbu2020monte} (MCMC) method, from the current model, and then the model parameters are updated based on the difference between the ``fake'' examples and the ``real'' observed examples in order to match the ``fake'' examples to the ``real'' observed examples in terms of some permutation-invariant statistical properties defined by the energy function.

Instead of implicitly modeling the distribution of points as a top-down generator \cite{goodfellow2014generative,kingma2013auto} 
(implicit because the marginal probability density of a generator model requires integrating out the latent noise vector, which is analytically  intractable) or indirectly learning the model by an adversarial learning scheme where a discriminator is recruited and simultaneously trained with the generator in a minimax two-player game, or a variational inference scheme where an encoder is used as an inference model to approximate the intractable posterior distribution, we explicitly model this distribution as an EBM and directly learn the model by MCMC-based MLE (as well as its variants) without the aid of any extra network. The MLE, in general, does not suffer from mode collapse and instability issues, which exist in GANs due to the unbalanced joint training of two models. 

Models using encoder-decoders for point cloud generation typically rely on hand-crafted distance metrics to measure the dissimilarity between two point sets. However, the MLE learning of our model corresponds to a statistical matching between the observed and the generated point clouds, where the statistical properties are defined by the derivatives of the energy function with respect to the learning parameters. Therefore, our model does not rely on hand-crafted distance metrics.

About the learning algorithm, as mentioned above, the MLE learning algorithm follows an ``analysis by synthesis'' scheme, which iterates the following two steps. Synthesis step: generate the ``fake'' synthesized examples from the current model. Analysis step: update the model parameters based on the difference between the ``real'' observed examples and the ``fake'' synthesized examples. See the recent paper \cite{nijkamp2019anatom} for a thorough investigation of various implementation schemes for learning the EBM.  The following are different implementations of the synthesis step. (i) Persistent chain \cite{xie2016theory}, which runs a finite-step MCMC such as Langevin dynamics \cite{neal2011mcmc} from the synthesized examples generated from the previous learning iteration. (ii) Contrastive divergence chain \cite{hinton2002training}, which runs a finite step MCMC from the observed examples. (iii) Non-persistent short-run MCMC \cite{nijkamp2019learning}, which runs a finite-step MCMC from Gaussian white noise. It is possible to learn an unbiased model using scheme (i), but the learning can be time-consuming. Scheme (ii) learns a biased model that usually cannot generate realistic synthesized examples. (iii) has been recently proposed by \cite{nijkamp2019learning}. Even though the learned model may still be biased, similar to contrastive divergence, the learning is very efficient, and the short-run MCMC initialized from noise can generate realistic synthesized examples. Moreover, the noise-initialized short-run Langevin dynamics may be viewed as a flow-like model  \cite{dinh2014nice,dinh2016density,kingma2018glow}  or a generator-like model \cite{goodfellow2014generative,kingma2013auto}  that transforms the initial noise to the synthesized example. Interestingly, the learned short-run dynamics is capable of reconstructing the observed examples and interpolating different examples, similar to the flow model and the generator model \cite{nijkamp2019learning}. 

In our work, we adopt the learning scheme (iii). We show that the learned short-run MCMC can generate realistic point cloud patterns, and it can reconstruct observed point clouds and interpolate between point clouds. Moreover, even though it learns a biased model, the learned energy function and features are still useful for classification.

\subsection{Related work}

\textbf{Energy-based modeling and learning}.  Energy-based generative ConvNets \cite{xie2016theory} aim to learn an explicit probability distribution of data in the form of the EBM, in which the energy function is parametrized by a modern convolutional neural network and the MCMC sampling is based on Langevin dynamics. Compelling results on learning complex data distributions with the energy-based generative ConvNets \cite{xie2016theory} have been shown on images \cite{xie2016theory}, videos \cite{xie2017synthesizing,xie2019synthesizing,han2019divergence} and 3D voxels \cite{xie2018learning,Xie2020GenerativeVL}. Some alternative sampling strategies to make the training of the models more effective have been studied. For example,  \cite{gao2018learning} proposes a multi-grid method for learning energy-based generative ConvNet models. 
Cooperative learning or CoopNets \cite{xie2016cooperative,xie2018cooperative,xie2021cooperative} trains a generative ConvNet with a generator as an amortized sampler via MCMC teaching. \cite{nijkamp2019learning} proposes to learn a non-convergent, non-mixing, and non-persistent short-run MCMC, and treats this short-run MCMC as a learned generator model. 
Recent advances show that the generative ConvNet can be trained with a VAE, e.g.,~\cite{han2019divergence, xie2020learning} or a flow-based model, e.g.,~\cite{gao2020flow,nijkamp2020learning}. However, the models in the works mentioned above are only suitable for data with regular structures. Learning EBMs for 3D point clouds, which are unordered point sets, has not been investigated prior to our paper.

\textbf{Deep learning for point clouds}. Deep learning methods have been successfully applied to point clouds for discriminative tasks including classification and segmentation, such as \cite{qi2017pointnet,qi2017pointnet++,zaheer2017deep}. 
PointNet \cite{qi2017pointnet} is a pioneering discriminative deep net that directly processes point clouds for classification, by designing permutation invariant network architecture to deal with unordered point sets. As to generative models of point clouds, 
\cite{gadelha2018multiresolution} uses VAEs and \cite{zamorski2018adversarial} uses adversarial auto-encoders with heuristic loss functions measuring the dissimilarity between two point sets, e.g., Chamfer distance (CD) or earth mover's distance (EMD), for the point cloud generation. GANs for point clouds are explored in \cite{li2018point,achlioptas2018learning,valsesia2018learning}. For example, \cite{li2018point} and \cite{achlioptas2018learning} learn a GAN on raw point cloud data, while \cite{li2018point} learns a GAN on the latent space of an auto-encoder that is pre-trained with CD or EMD loss on raw data. \cite{valsesia2018learning} proposes to generate point clouds via a GAN with graph convolution that extracts localized information from point clouds. \cite{yang2019pointflow} studies point cloud generation using continuous normalizing flows trained with variational inference. Our paper learns an EBM of point clouds via MCMC-based MLE.  
The proposed model, which we call \textit{Generative PointNet} (or GPointNet), can be derived from the discriminative PointNet. Our model enables us to get around the complexities of training GANs or VAEs, or the troubles of crafting distance metrics for measuring similarity between two point sets.

\subsection{Contributions}
The key contributions of our work are as follows.

\textbf{Modeling}: We propose a novel EBM to explicitly represent the probability distribution of an unordered point set, e.g., a 3D point cloud, by designing a input-permutation-invariant bottom-up network as the energy function. This is the first generative model that provides an explicit density function for point cloud data. It will shed a new light not only on the area of 3D deep learning but also in the study of unordered set modeling.

\textbf{Learning}: Under the proposed EBMs, we propose to adopt an unconventional short-run MCMC to learn our model and treat the MCMC as a flow-based generator model, such that it can be used for point cloud reconstruction and generation simultaneously. Usually EBM is unable to reconstruct data. This is the first EBM that can perform point cloud reconstruction and interpolation.

\textbf{Uniqueness}: Compared with  existing point cloud generative models, our model has the following unique properties: (1) It does not rely on an extra assisting network for training; (2) It can be derived from the discriminative PointNet; (3) It unifies synthesis and reconstruction in a single framework; (4) It unifies an explicit density (i.e., EBM) and an implicit density (i.e., short-run MCMC as a latent variable model) of the point cloud in a single framework.   

\textbf{Performance}: Our energy-based framework obtains competitive performance with much fewer parameters compared with the state-of-art point cloud generative models, such as GAN-based and VAE-based approaches, in the tasks of synthesis, reconstruction and classification.

\section{Generative PointNet}

\subsection{Energy-based model for unordered point sets} 
Suppose we observe a set of 3D shapes $\{X_i, i=1,...,N\}$ from a particular category of objects. Each shape is represented by a set of 3D points $X=\{x_k, k=1,...,M\}$, where each point $x$ is a vector of its 3D coordinate plus optional extra information such as RGB color, etc. In this paper, the points we discuss only contain 3D coordinate information for simplicity. 

We define an explicit probability distribution of shape, each shape itself being a 3D point cloud, by the following energy-based model 
\begin{eqnarray}
p_\theta(X) = \frac{1}{Z(\theta)} \exp\left[ f_\theta(X) \right]p_0(X), 
\label{eq:model}
\end{eqnarray} 
where $f_\theta(X)$ is a scoring function that maps the input $X$ to a score and is parameterized by a bottom-up neural network, $p_0(X) \propto \exp(-||X||^2/2s^2)$ is the Gaussian white noise reference distribution ($s$ is a hyperparameter and set to be 0.3 in our paper), $Z(\theta)=\int \exp[f_{\theta}(X)]p_0(X)dX$ is the analytically intractable normalizing constant, which ensures the sum of all the probabilities in the distribution is equal to~1. The energy function $\mathcal{E}_{\theta}(X)=-f_\theta(X)+||X||^2/2s^2$ containing parameters $\theta$ defines the energy of the point cloud $X$, and the point cloud $X$ with a low energy is assigned a high probability. 

\begin{figure*}[h]
\centering	
\includegraphics[width=1\linewidth]{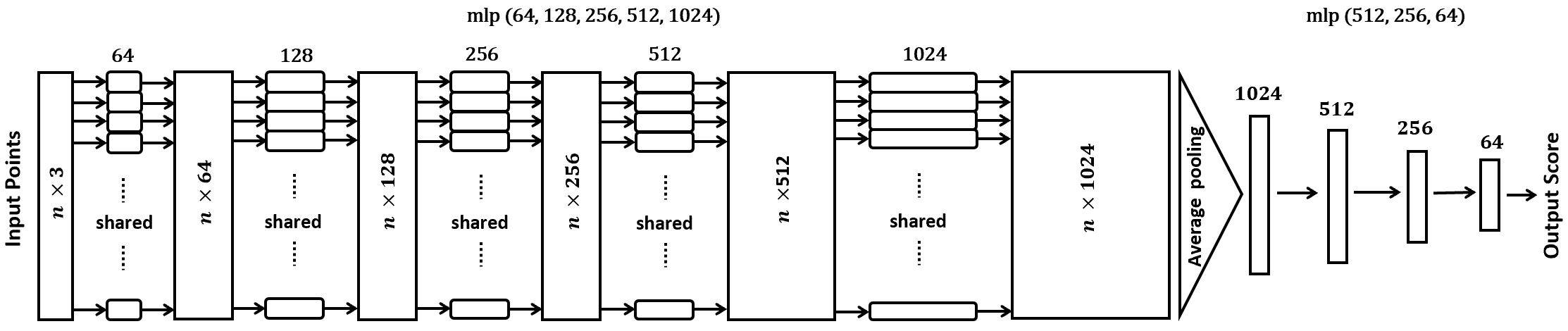}
\caption{Architecture of the scoring function of the Generative PointNet. The scoring function $f_{\theta}(X)$ is an input-permutation-invariant bottom-up deep network, which takes $n$ unordered points as input, encodes each point into features by multilayer perceptron (MLP) with numbers of channels 64, 128, 256, 512 and 1,024 at each layer respectively, and then aggregates all point features to a global feature by average pooling, and eventually outputs scalar energy by multilayer perceptron with numbers of channels 512, 256, 64 and 1 at each layer respectively. Layer Normalization \cite{ba2016layer} is used with ReLU for layers before average pooling, while only ReLU is used for layers after average pooling.}
\label{fig:structure}
\end{figure*}

Since each point cloud input $X$ is a set of unordered points, the energy function, $\mathcal{E}_\theta(X)$, defined on a point set needs to be invariant to $M!$ permutations of the point set in point feeding order. Because $||X||^2/2s^2$ is already naturally invariant to the point permutation, we only need to parameterize $f_\theta(X)$ by an input-permutation-invariant bottom-up deep network in order to obtain a proper  $\mathcal{E}_\theta(X)$ that can handle unordered points. Specifically, we design $f_\theta(X)$ by applying a symmetric function on non-linearly transformed points in the set, i.e., $f_{\theta}(\{x_1,...,x_M\})=g(\{h(x_1),..,h(x_M)\})$, 
where $h$ is parameterized by a multi-layer perceptron network and $g$ is a symmetric function, which is an average pooling function followed by a multi-layer perceptron network. The network architecture of the scoring function $f_{\theta}$ is illustrated in Figure \ref{fig:structure}. Please read the caption for the details of the network. 

\subsection{Maximum likelihood} 
Suppose we observe a collection of 3D point clouds $\mathcal{X}=\{X_i, i=1,...,N\}$ from a particular category of object. Let $q_{\rm data}$ be the distribution that generates the observed examples. The goal of learning  $p_{\theta}$ is to estimate the parameter $\theta$ from the observations $\mathcal{X}$. For a large $N$, the maximum likelihood estimation of $\theta$,
\[ 
\max_{\theta} \left[ \frac{1}{N} \sum_{i=1}^{N}\log p_{\theta}(X_i)\right] \approx \max_{\theta} \E_{q_{\rm data}} \left[ \log p_{\theta}(X)\right] \nonumber
\]
is equivalently to minimize the Kullback-Leibler (KL)-divergence ${\rm KL}(q_{\rm data}\|p_\theta)$ over $\theta$, where the ${\rm KL}$ divergence is defined as ${\rm KL}(q|p)=\E_{q}[\log(q(x)/p(x))]$. We can update $\theta$ by gradient ascent. The gradient of the log-likelihood or, equivalently, the negative KL divergence is computed by
\begin{eqnarray} 
 &&  - \frac{\partial}{\partial \theta} {\rm KL}(q_{\rm data}(X)\|p_\theta(X)) \nonumber\\ 
  &= & \E_{q_{\rm data}} \left[\frac{\partial}{\partial \theta} f_\theta(X)\right] -  \E_{p_\theta} \left[\frac{\partial}{\partial \theta} f_\theta(X)\right] \label{eq:gradient}\\
  &\approx & \frac{1}{n} \sum_{i=1}^{n}\left[\frac{\partial}{\partial \theta} f_\theta(X_i)\right]  -  \frac{1}{n} \sum_{i=1}^{n}\left[\frac{\partial}{\partial \theta} f_\theta(\tilde{X}_i)\right], \label{eq:gradient_MCMC}
\end{eqnarray} 
where $\{\tilde{X}_i,i=1,...,n\}$ are $n$ point clouds generated from the current distribution $p_{\theta}$ by an MCMC method, such as Langevin dynamics. Eq.(\ref{eq:gradient_MCMC}) refers to the MCMC approximation of the analytically intractable gradient due to the intractable expectation term $\E_{p_{\theta}}[\cdot]$ in Eq.(\ref{eq:gradient}), and leads to the mini-batch ``analysis by synthesis'' learning algorithm. At iteration $t$, we randomly sample a batch of observed examples from the training data set $\{X_i,i=1,...,n\} \sim q_{\rm data}$, and generate a batch of synthesized examples from the current distribution $\{\tilde{X}_i,i=1,...,n\} \sim p_{\theta}$ by MCMC sampling. Then we compute the gradient $\Delta(\theta_t)$ according to Eq.(\ref{eq:gradient_MCMC}) and update the model parameter $\theta$ by $\theta_{t+1}=\theta_{t}+ \gamma_t \Delta(\theta_t)$ with a learning rate $\gamma_{t}$.

\subsection{MCMC sampling with Langevin dynamics}
To sample point clouds from the distribution $p_{\theta}(X)$ by Langevin dynamics, we iterate the following step:
\begin{eqnarray} 
 && X_{\tau+ 1} = X_{\tau} - \frac{\delta^2 }{2} \frac{\partial}{\partial X} \mathcal{E}_{\theta}(X_{\tau})  + \delta U_{\tau},  \label{eq:MCMC}
\end{eqnarray} 
where $\tau$ indexes the time step, $\delta$ is the step size, and $U_{\tau} \sim {\mathcal N}(0,I)$ is the Gaussian white noise. Since $f_{\theta}$ is a differentiable function, the term of gradient of $\mathcal{E}_{\theta}(X_{\tau})$ with respect to $X$ can be efficiently computed via back-propagation. 
As to MCMC initialization, the following are three options. (1) Initialize long-run non-persistent MCMC from noise point clouds. (2) Initialize persistent MCMC from noise point clouds, and within each subsequent learning iteration, run a finite-step MCMC starting from the synthesized point cloud generated in the previous learning iteration. (3) Following Contrastive Divergence \cite{hinton2002training}, one may initialize the MCMC from the training examples sampled from the training data set within each learning iteration.  

\begin{figure*}[!h]
	\centering	
    \setlength{\tabcolsep}{1pt}
    \renewcommand{\arraystretch}{0.5}
    \begin{tabular}{cccc@{\hskip 2mm}|@{\hskip 2mm}ccccccccc} 
    	\hspace{2mm}\rotatebox{90}{\hspace{3mm}{\small chair}} &
        \includegraphics[trim=30 30 30 30,clip,width=0.07\linewidth]{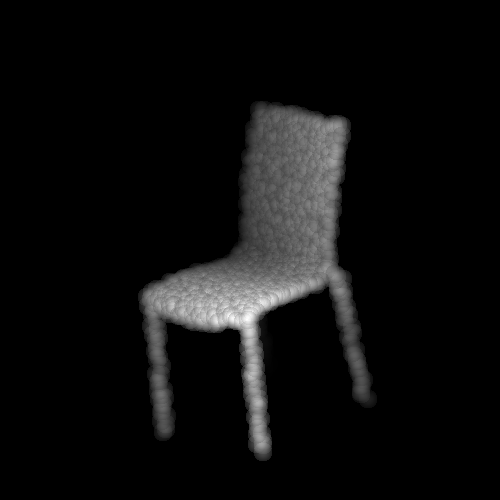} &
        \includegraphics[trim=30 30 30 30,clip,width=0.07\linewidth]{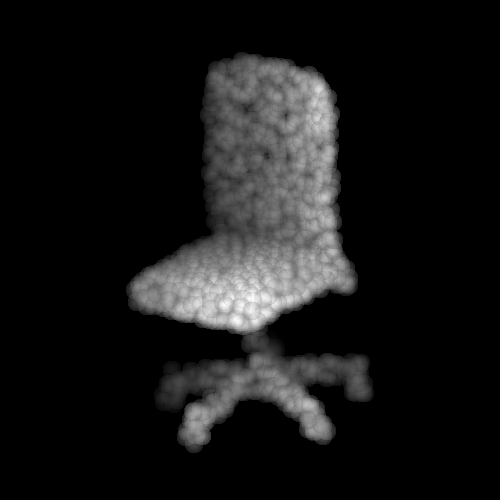} &
        \includegraphics[trim=30 30 30 30,clip,width=0.07\linewidth]{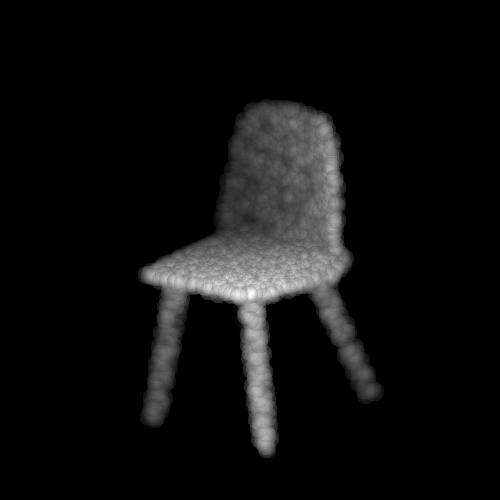} &
        \includegraphics[trim=30 30 30 30,clip,width=0.07\linewidth]{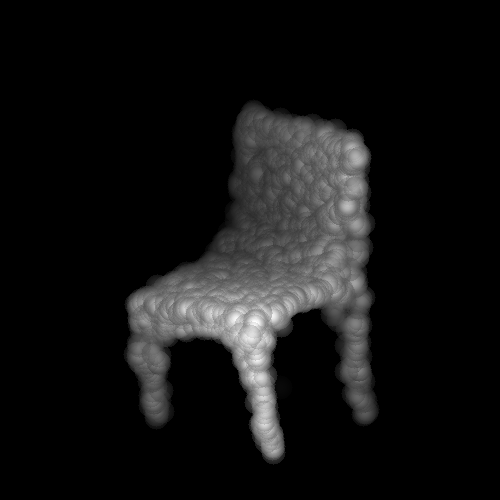} &
        \includegraphics[trim=30 30 30 30,clip,width=0.07\linewidth]{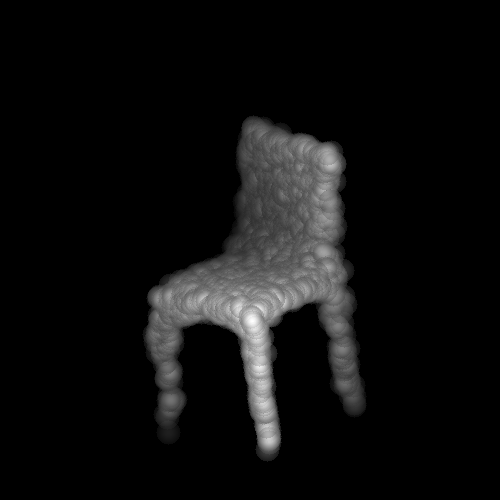} &
        \includegraphics[trim=30 30 30 30,clip,width=0.07\linewidth]{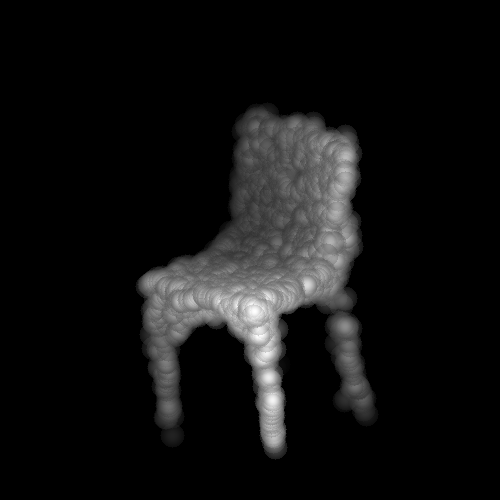} &
        \includegraphics[trim=30 30 30 30,clip,width=0.07\linewidth]{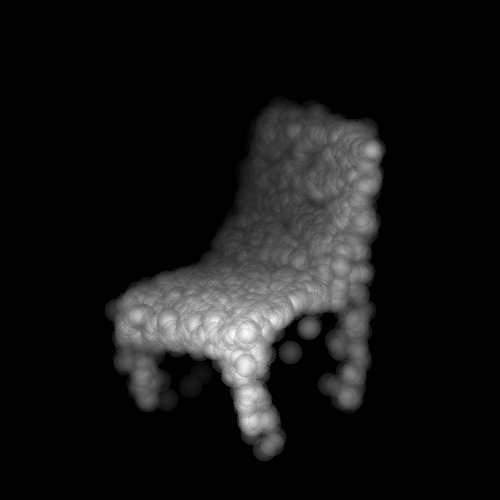} &
        \includegraphics[trim=30 30 30 30,clip,width=0.07\linewidth]{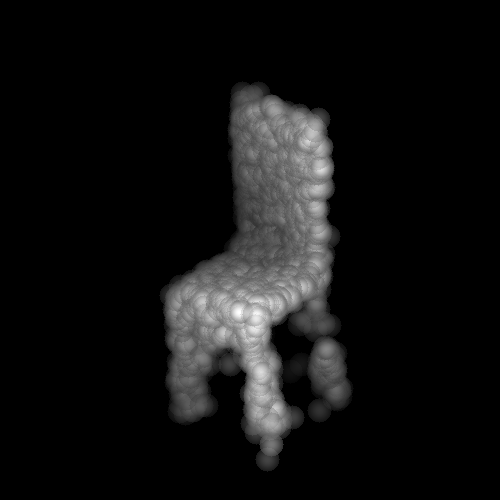} &
        \includegraphics[trim=30 30 30 30,clip,width=0.07\linewidth]{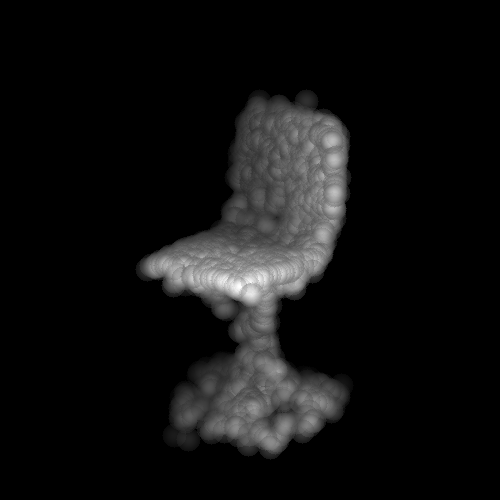} &
        \includegraphics[trim=30 30 30 30,clip,width=0.07\linewidth]{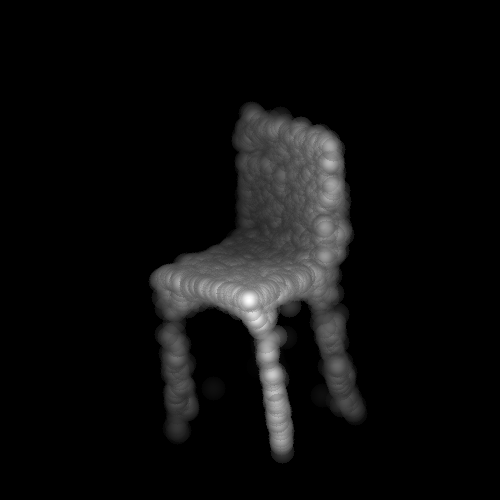} &
        \includegraphics[trim=30 30 30 30,clip,width=0.07\linewidth]{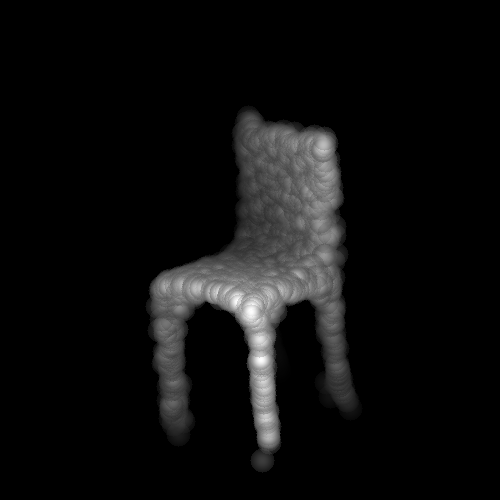} &
        \includegraphics[trim=30 30 30 30,clip,width=0.07\linewidth]{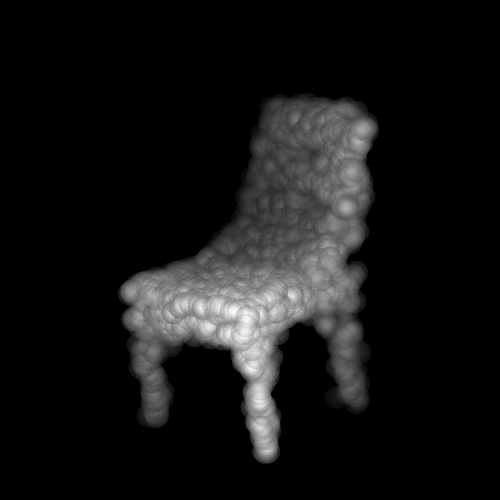} 
        \\
        \hspace{2mm}\rotatebox{90}{\hspace{3mm}{\small toilet}} &
        \includegraphics[trim=30 30 30 30,clip,width=0.07\linewidth]{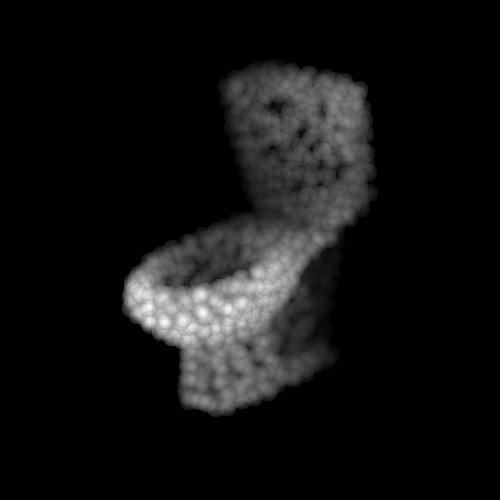} &
        \includegraphics[trim=30 30 30 30,clip,width=0.07\linewidth]{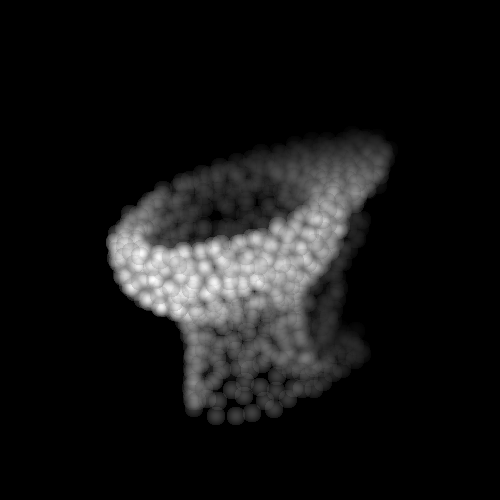} &
        \includegraphics[trim=30 30 30 30,clip,width=0.07\linewidth]{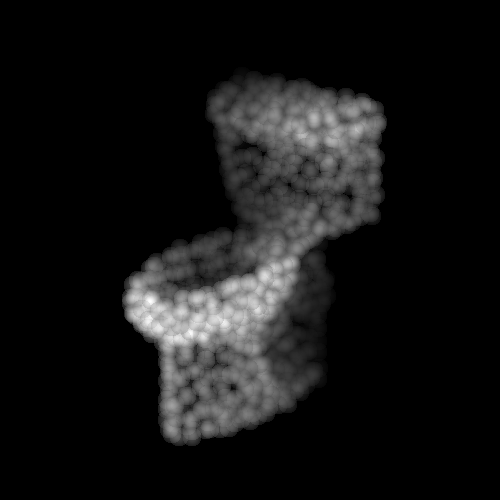} &
        \includegraphics[trim=30 30 30 30,clip,width=0.07\linewidth]{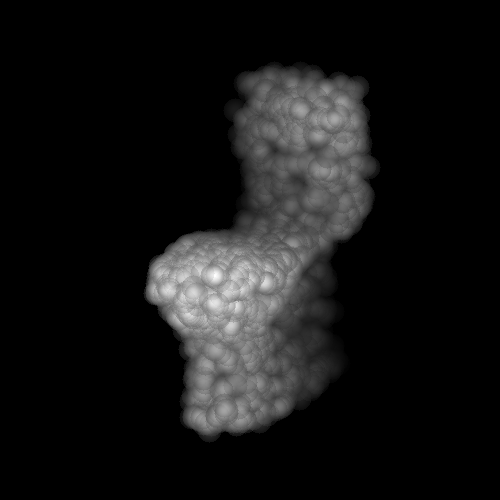} &
        \includegraphics[trim=30 30 30 30,clip,width=0.07\linewidth]{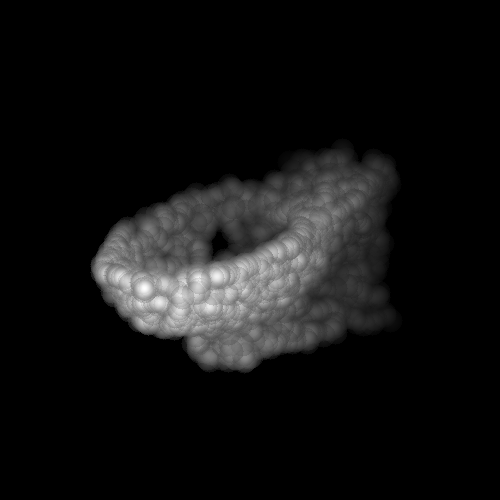} &
        \includegraphics[trim=30 30 30 30,clip,width=0.07\linewidth]{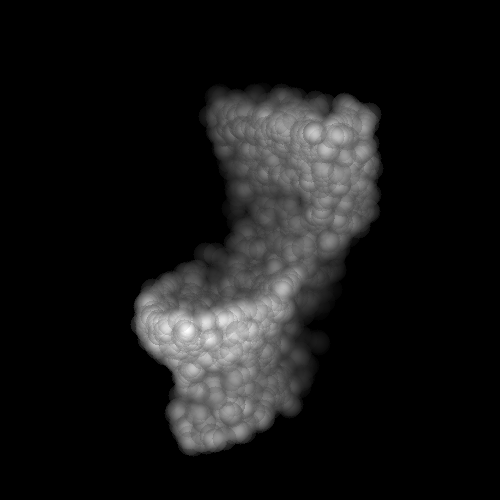} &
        \includegraphics[trim=30 50 30 10,clip,width=0.07\linewidth]{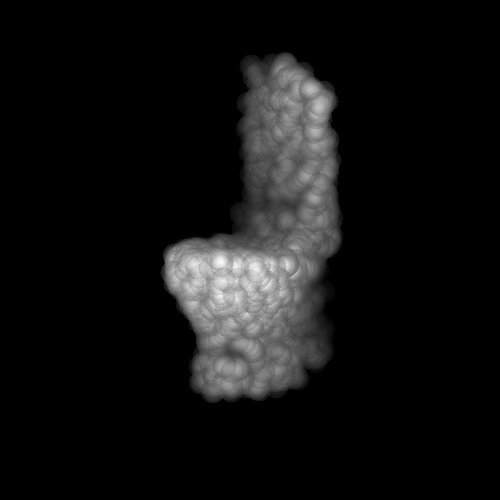} &
        \includegraphics[trim=30 30 30 30,clip,width=0.07\linewidth]{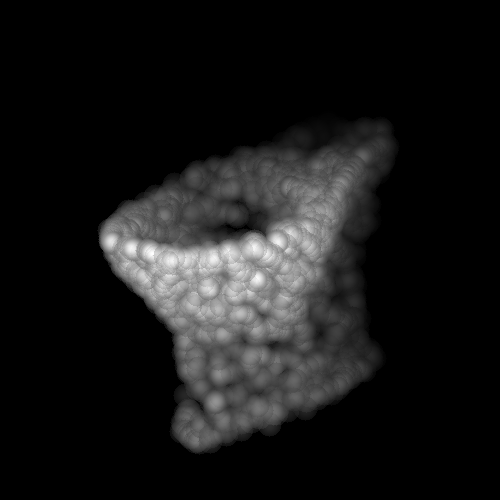} &
        \includegraphics[trim=30 50 30 10,clip,width=0.07\linewidth]{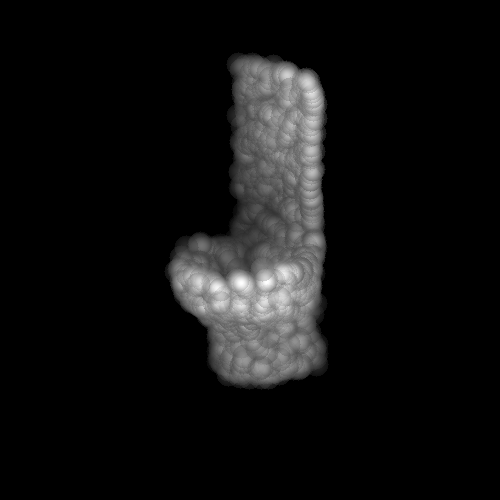} &
        \includegraphics[trim=30 30 30 30,clip,width=0.07\linewidth]{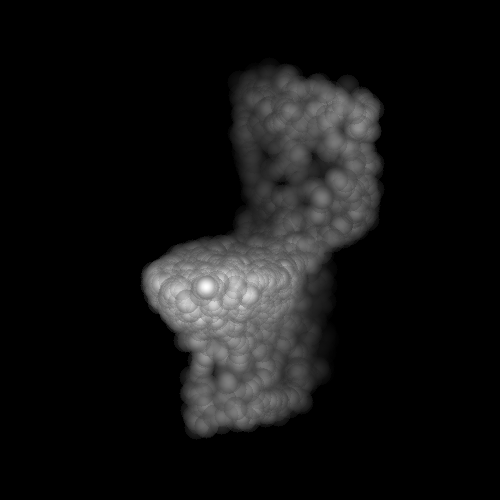} &
        \includegraphics[trim=30 30 30 30,clip,width=0.07\linewidth]{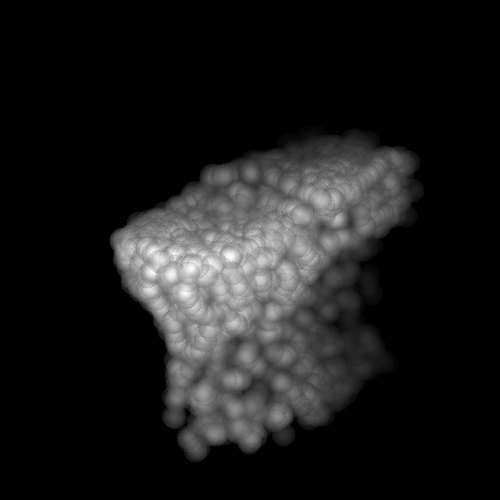} &
        \includegraphics[trim=30 30 30 30,clip,width=0.07\linewidth]{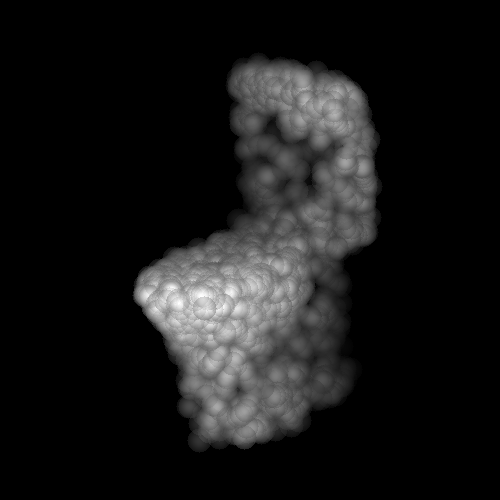} 
        \\ 
        \hspace{2mm}\rotatebox{90}{\hspace{3mm}{\small table}} &      
        \includegraphics[trim=50 30 50 70,clip,width=0.07\linewidth]{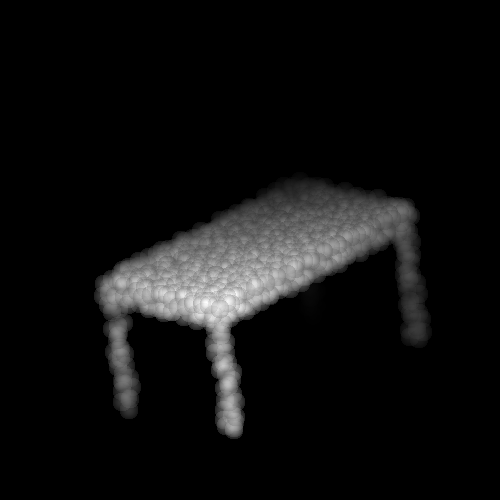} &
        \includegraphics[trim=50 30 50 70,clip,width=0.07\linewidth]{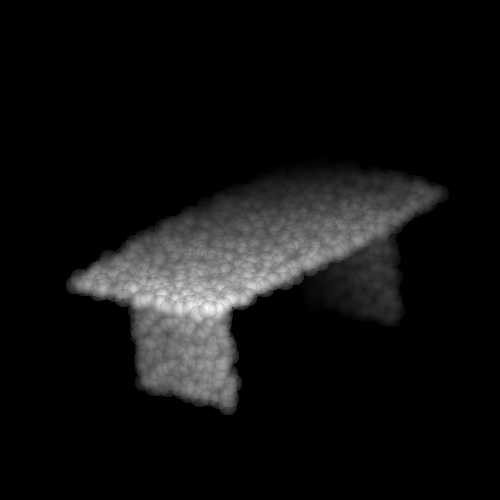} &
        \includegraphics[trim=50 30 50 70,clip,width=0.07\linewidth]{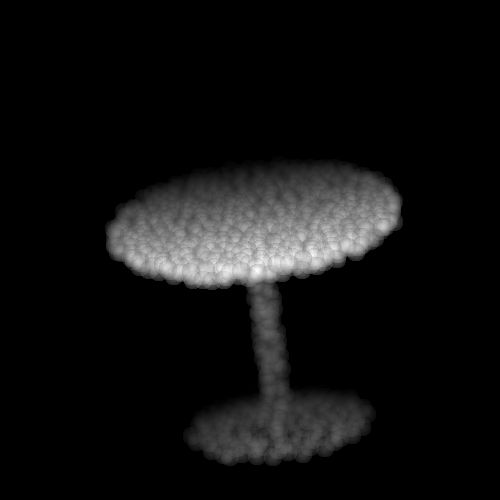} &
        \includegraphics[trim=30 10 30 50,clip,width=0.07\linewidth]{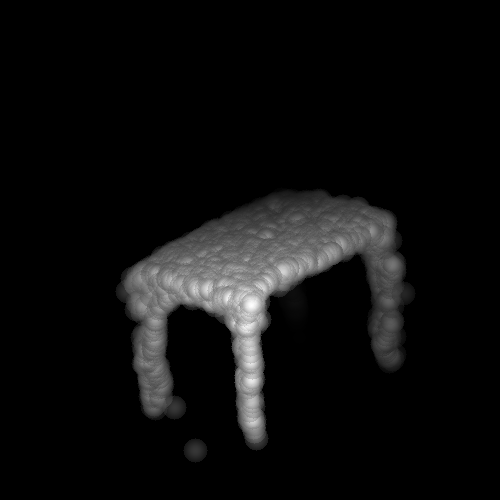} &
        \includegraphics[trim=30 10 30 50,clip,width=0.07\linewidth]{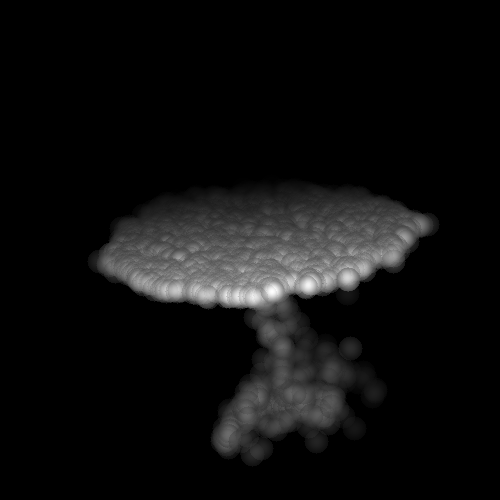} &
        \includegraphics[trim=30 10 30 50,clip,width=0.07\linewidth]{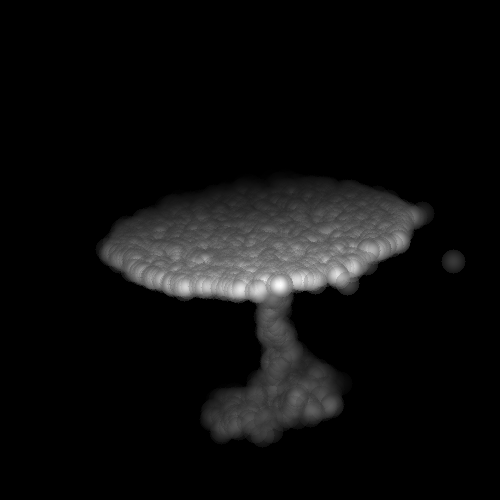} &
        \includegraphics[trim=30 10 30 50,clip,width=0.07\linewidth]{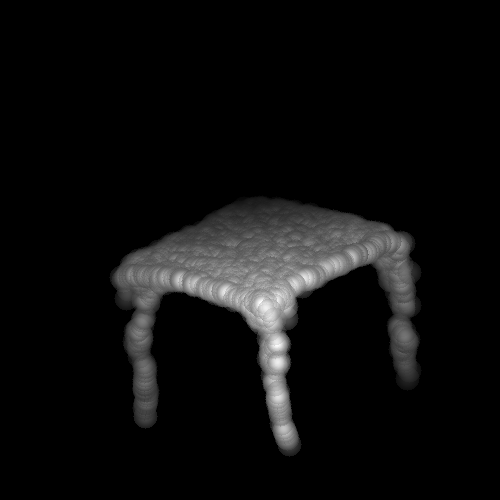} &
        \includegraphics[trim=30 10 30 50,clip,width=0.07\linewidth]{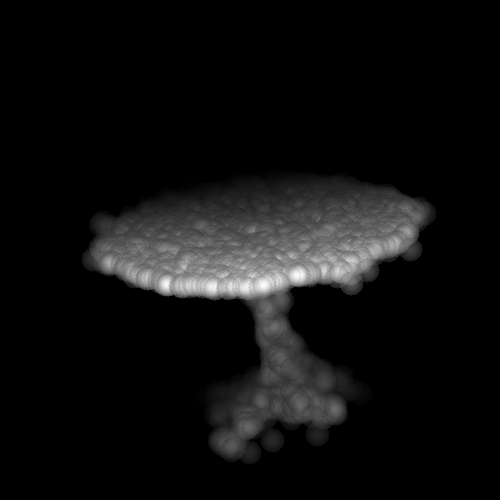} &
        \includegraphics[trim=30 10 30 50,clip,width=0.07\linewidth]{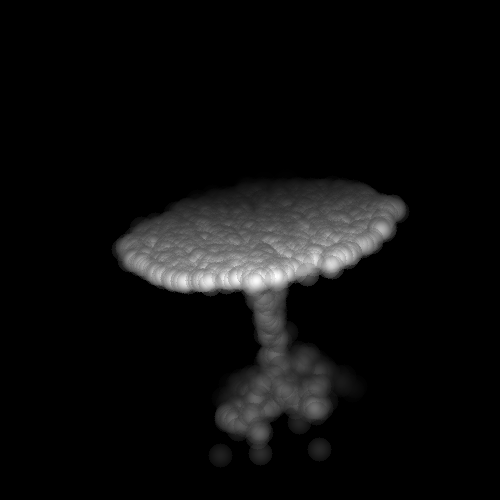} &
        \includegraphics[trim=30 10 30 50,clip,width=0.07\linewidth]{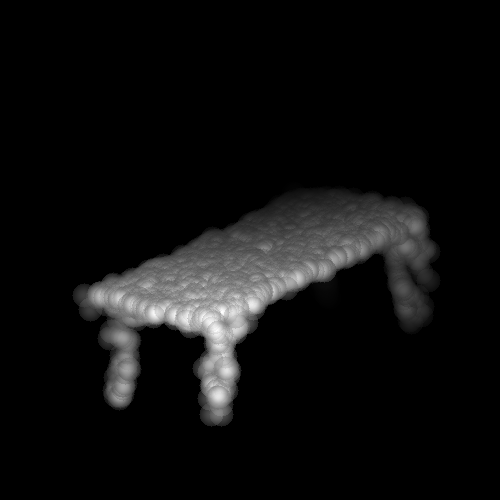} &
        \includegraphics[trim=30 10 30 50,clip,width=0.07\linewidth]{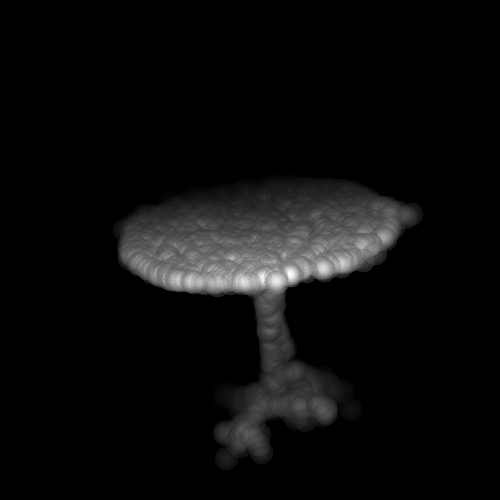} &
        \includegraphics[trim=30 10 30 50,clip,width=0.07\linewidth]{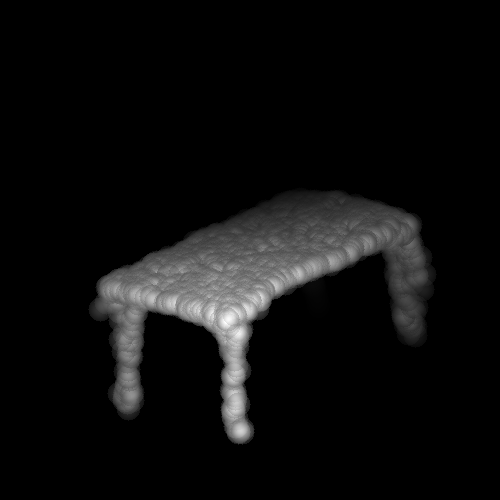} 
        \\
        \hspace{2mm}\rotatebox{90}{\hspace{1.9mm}{\small bathtub}} &
        \includegraphics[trim=30 30 30 30,clip,width=0.07\linewidth]{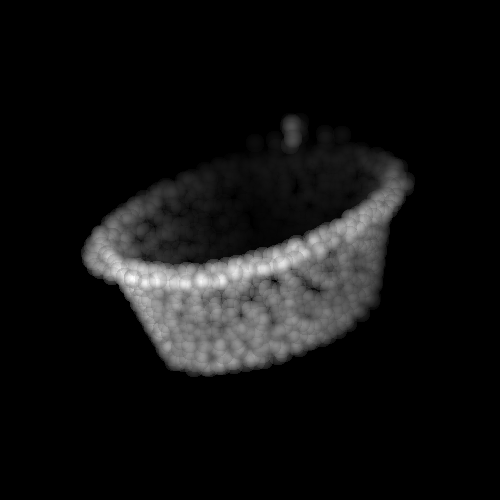} &
        \includegraphics[trim=30 30 30 30,clip,width=0.07\linewidth]{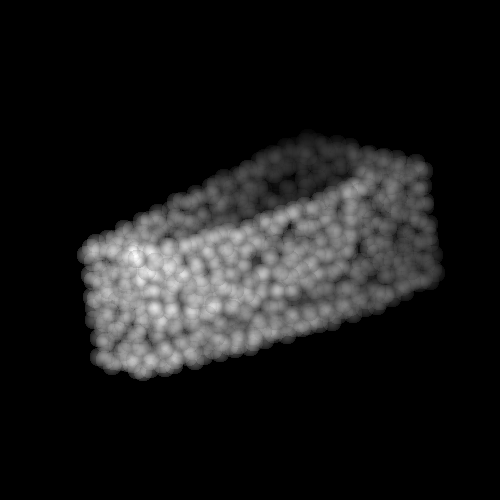} &
        \includegraphics[trim=30 30 30 30,clip,width=0.07\linewidth]{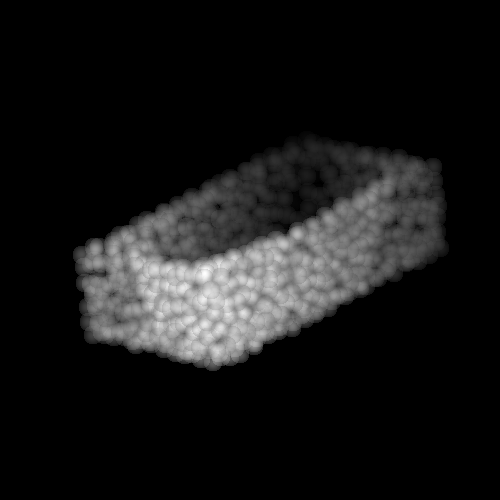} &
        \includegraphics[trim=30 30 30 30,clip,width=0.07\linewidth]{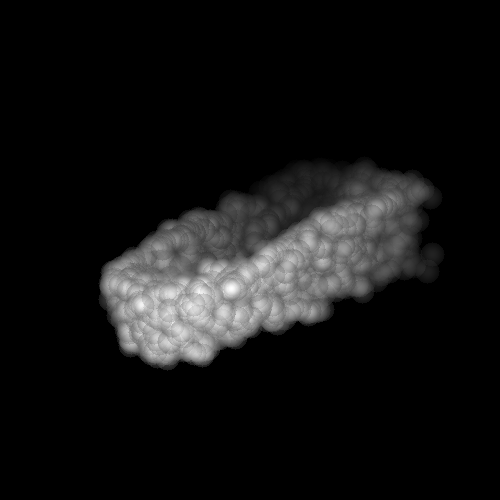} &
        \includegraphics[trim=30 30 30 30,clip,width=0.07\linewidth]{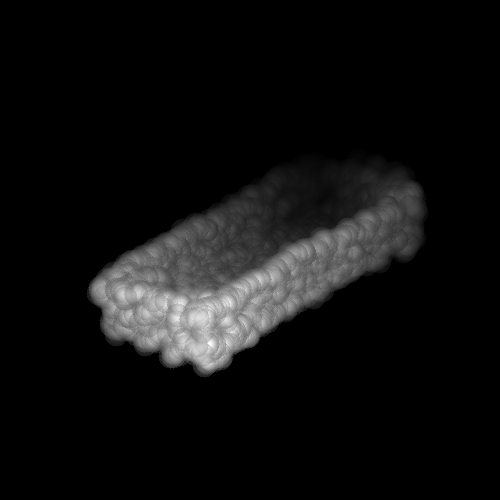} &
        \includegraphics[trim=30 30 30 30,clip,width=0.07\linewidth]{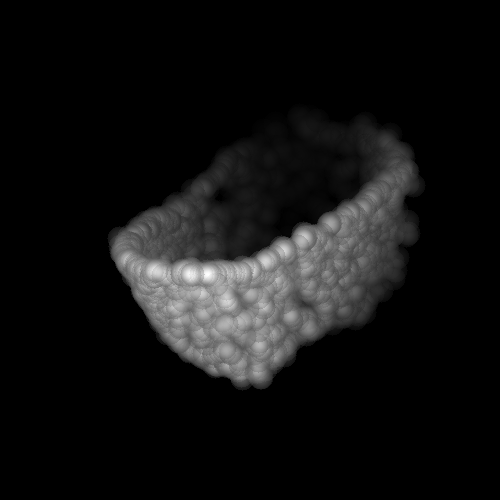} &
        \includegraphics[trim=30 30 30 30,clip,width=0.07\linewidth]{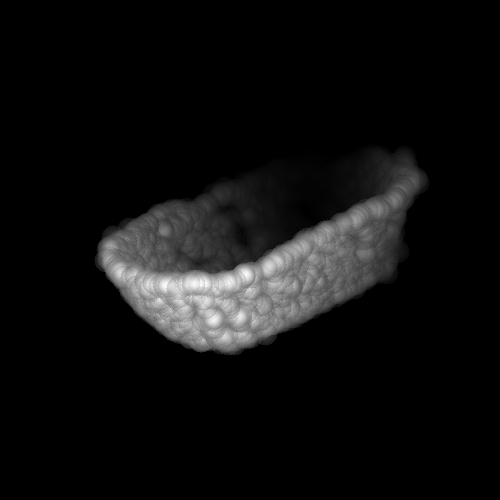} &
        \includegraphics[trim=30 30 30 30,clip,width=0.07\linewidth]{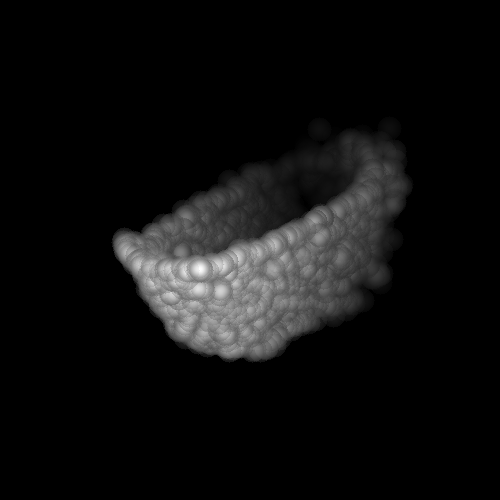} &
        \includegraphics[trim=30 30 30 30,clip,width=0.07\linewidth]{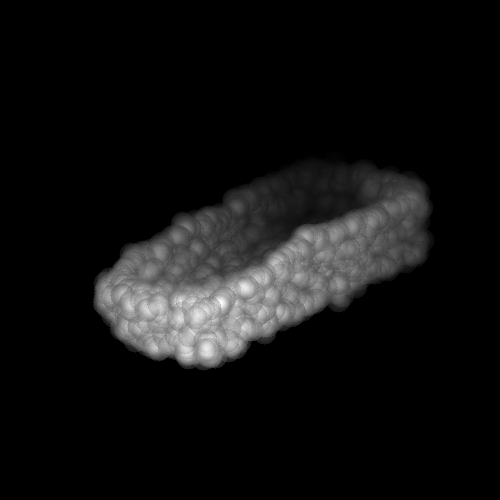} &
        \includegraphics[trim=30 30 30 30,clip,width=0.07\linewidth]{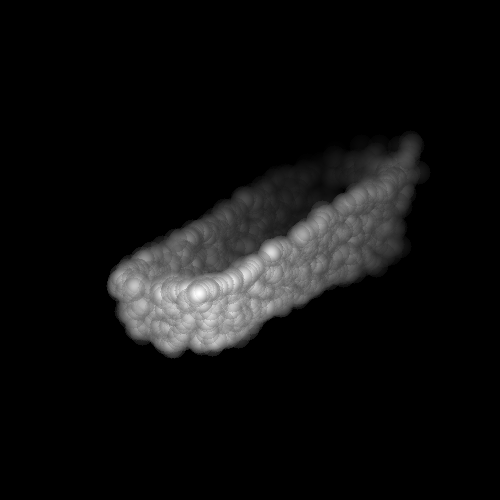} &
        \includegraphics[trim=30 30 30 30,clip,width=0.07\linewidth]{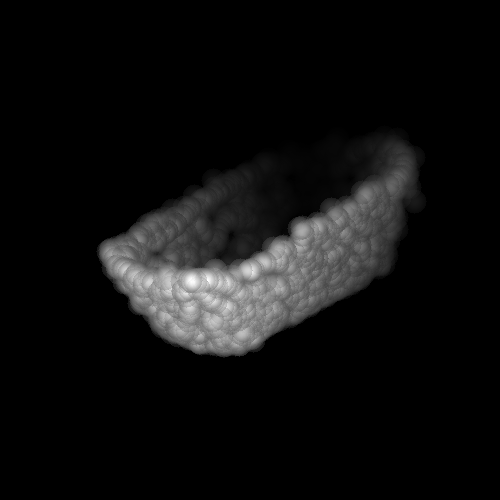} &
        \includegraphics[trim=30 30 30 30,clip,width=0.07\linewidth]{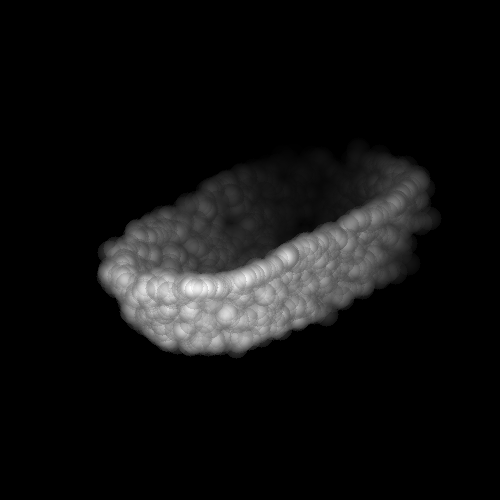}
    \end{tabular}
	\caption{Generating 3D point clouds of objects. Each row shows one experiment, where the first three point clouds are three examples randomly selected from the training set. The rest are synthesized point clouds sampled from the short-run Langevin dynamics. The number of points in each example is 2,048. From top to bottom: chair, toilet, table, and bathtub.}
	\label{fig:syn}
\end{figure*}
\section{Short-run MCMC as generator model}
\label{sec:generator}
Learning $p_{\theta}$ requires MCMC sampling to generate synthesized point clouds. The learned $p_{\theta}$ is multi-modal because $p_{\text data}$ is usually multi-modal, which is due to the complexity of the point cloud patterns and the large scale of the data set. The property of multimodality is likely to cause different MCMC chains to get trapped by the local modes. Thus the MCMC sampling of $p_{\theta}$ may take a long time to mix, regardless of the initial distribution and the length of the Markov chain. Following the recent work on learning EBM \cite{nijkamp2019learning}, instead of running a long-run convergent MCMC to sample from $p_{\theta}$, we only run non-convergent, non-persistent short-run MCMC  toward $p_{\theta}$ for a fixed number of steps $K$, starting from a fixed initial distribution, such as Gaussian white noise distribution $p_0$. 

We use $M_{\theta}$ to denote the transition kernel of the $K$ steps of MCMC toward $p_{\theta}(X)$. For a given initial probability distribution $p_0$, the resulting marginal distribution of the sample $X$ after running $K$ steps of MCMC starting from $p_0$ is denoted by
\begin{eqnarray} 
 q_{\theta}(X)=M_{\theta}p_{0}(X)= \int p_{0}(Z)M_{\theta}(X|Z)dZ \label{eq:short_run_MCMC}
\end{eqnarray} 

Since $q_{\theta}(X)$ is not convergent, the $X$ is highly dependent to $Z$. $q_{\theta}(X)$ can be considered a generator model, a flow-based model, or a latent variable model with $Z$ being the continuous latent variables in the following form
\begin{equation}
 X = M_{\theta}(Z,\xi), \hspace{1mm} Z \sim p_{0}(Z),\label{eq:generator}
\end{equation} 
where $Z$ and $X$ have the same number of dimensions, and $Z$ follows a known prior  (Gaussian) distribution $p_0$. $M_{\theta}$ is a short-run Langevin dynamics including $K$ Langevin steps in Eq.(\ref{eq:MCMC}), which can be considered a $K$-layer residual network with noise injected into each layer and weight sharing at each layer. Let $\xi$ be all the randomness in $M_{\theta}$ due to the layer-wise injected noise. 
The model represented by a short-run MCMC shown in Eq.(\ref{eq:generator}) can be trained by the ``analysis by synthesis'' scheme, where we update $\theta$ according to Eq.(\ref{eq:gradient_MCMC}) and synthesize $\{\tilde X\}$ according to Eq.(\ref{eq:generator}).   Training $\theta$ with a short-run MCMC is no longer a maximum likelihood estimator but a moment matching estimator (MME) that solves the following estimating equation 
\begin{eqnarray} 
\E_{q_{\rm data}} \left[\frac{\partial}{\partial \theta} f_\theta(X)\right] =  \E_{p_\theta} \left[\frac{\partial}{\partial \theta} f_\theta(X)\right].    \label{eq:short_run_eq}
\end{eqnarray} 
Even though the learned $p_{\theta}$ based on short-run MCMC is $p_{{\hat{ \theta}}_{\text{ MEE}}}$ rather than $p_{{\hat{ \theta}}_{\text{ MLE}}}$, the $q_{{\hat{ \theta}}_{\text{ MEE}}}$ is still a valid generator that is useful for 3D point cloud generation and reconstruction. As to reconstruction, given a testing 3D point cloud $X$, we can reconstruct $X$ by finding $Z$ to minimize the reconstruction error $L(Z)=\|X - M_{\theta}(Z)\|^2$, where $M_{\theta}(Z)$ is a noise-disabled version of $M_{\theta}(Z, \xi)$ (after learning, the noise term is negligible compared to the gradient term). This can be easily achieved by running gradient descent on $L(Z)$, with $Z$ initialized from $Z_0 \sim p_0$. Even though we abandon $p_{\theta}$ in Eq.(\ref{eq:model}) and keep $q_{\theta}$ in Eq.(\ref{eq:short_run_MCMC}) eventually, $p_{\theta}$ is crucial because $q$ is derived from $p$ and we learn $q$ under $p$. In other works, $p$ serves as an incubator of $q_{{\hat{ \theta}}_{\text{ MEE}}}$.

When the model $p_{\theta}$ is learned from a large scale data set and only a limited budge of MCMC can be affordable, learning a short-run MCMC as a generator model toward $p_{\theta}$ for point cloud generation and construction will be a tradeoff between MCMC efficiency and MLE accuracy.  

The learning method based on noise-initialized short-run MCMC is similar to contrastive divergence \cite{hinton2002training}, which initializes a finite-step MCMC from each observed example within each learning iteration. Contrastive divergence also learns a bias model, but the learned model is usually incapable of synthesis, much less reconstruction and interpolation. For noise-initialized short-run Langevin, it is possible to optimize tuning parameters such as step size $\delta$ to minimize the bias caused by short-run MCMC. 
Also, the learning algorithm of our model seeks to match the expectations of $\Phi_\theta(X) = \frac{\partial}{\partial \theta} f_\theta(X)$ over the observed data and synthesized data. In the recent literature on the theoretical understanding of deep neural networks, the expectation of $\langle \Phi_\theta(X), \Phi_\theta(X')\rangle$, where the expectation is with respect to the random initialization of $\theta$, is called the neural tangent kernel \cite{jacot2018neural}, and it plays a central role in understanding the optimization and generalization of deep and wide networks. It is possible to define a metric based on such a kernel. We shall study these issues in our future work.

\begin{table*}[t]
	\small
	\centering
	\begin{tabular}{|l|l|c|cc|cc|}
		\hline
		\multirow{2}{*}{\rotatebox{90}{}}  & \multirow{2}{*}{Model}   &  \multirow{2}{*}{JSD ($\downarrow$)}  & \multicolumn{2}{c|}{MMD ($\downarrow$)}        & \multicolumn{2}{c|}{Coverage ($\uparrow$)}    \\ \cline{4-7} 
		&                       &             & CD            & EMD            & CD             & EMD            \\\hline \hline

        \multirow{5}{*}{\rotatebox{90}{night stand}} & r-GAN        & 2.679            & 1.163            & 2.394            & 50.00          & 38.37          \\
                                    & l-GAN        & 1.000            & 0.746            & 1.563            & 44.19          & 39.53          \\
                                    & PointFlow    & \textbf{0.240} & 0.888 & 1.451 & 55.81 & 39.53             \\
                                    & Ours         & 0.590            & \textbf{0.692}   & \textbf{1.148}   & \textbf{59.30} & \textbf{61.63} \\\cline{2-7}
                                    & Training Set & 0.263          & 0.793          & 1.096          & 60.40           & 52.32          \\\hline
        \multirow{5}{*}{\rotatebox{90}{toilet}}       & r-GAN        & 3.180            & 2.995            & 2.891            & 17.00          & 16.00             \\
                                    & l-GAN        & 1.253            & 1.258            & 1.481            & 21.00          & 28.00             \\
                                    & PointFlow    & \textbf{0.362} & 0.965 & 1.513 & 39.00 & 33.00             \\
                                    & Ours         & 0.386          & \textbf{0.816} & \textbf{1.265} & \textbf{44.00}    & \textbf{37.00}    \\\cline{2-7}
                                    & Training Set & 0.249          & 0.823          & 1.116           & 48.00         & 51.00          \\\hline
        \multirow{5}{*}{\rotatebox{90}{monitor}}      & r-GAN        & 2.936          & 1.524          & 2.021          & 21.00             & 24.00             \\
                                    & l-GAN        & 1.653          & 0.915          & 1.349          & 28.00             & 27.00             \\
                                    & PointFlow    & \textbf{0.326} & 0.831 & 1.288 & 37.00 & 32.00           \\
                                    & Ours         & 0.780            & \textbf{0.803} & \textbf{1.213} & \textbf{40.00}             & \textbf{38.00}    \\\cline{2-7}
                                    & Training Set & 0.283          & 0.554          & 0.938           & 48.00         & 53.00          \\\hline
        \multirow{5}{*}{\rotatebox{90}{chair}}        & r-GAN        & 2.772          & 1.709          & 2.164          & 23.00             & 28.00             \\
                                    & l-GAN        & 1.358          & 1.419          & 1.480            & 23.00             & 26.00             \\
                                    & PointFlow    & \textbf{0.278} & 0.965 & 1.322 & 42.00 & 51.00          \\
                                    & Ours         & 0.563          & \textbf{0.889} & \textbf{1.280} & \textbf{56.00}    & \textbf{57.00}    \\\cline{2-7}
                                    & Training Set & 0.365          & 0.858          & 1.190          & 54.00             & 59.00          \\\hline
        \multirow{5}{*}{\rotatebox{90}{bathtub}}      & r-GAN        & 3.014          & 2.478          & 2.536          & 26.00             & 30.00             \\
                                    & l-GAN        & 0.928          & 0.865          & 1.324           & 32.00             & 38.00             \\
                                    & PointFlow    & \textbf{0.350} & \textbf{0.593} & 1.320 & 50.00 & 44.00             \\
                                    & Ours         & 0.460          & 0.660          & \textbf{1.108} & \textbf{58.00} & \textbf{50.00}    \\\cline{2-7}
                                    & Training Set & 0.344          & 0.652           & 0.980          & 56.00             & 52.00          \\\hline
	\end{tabular}
	\hspace{1mm}
	\begin{tabular}{|l|l|c|cc|cc|}
		\hline
		\multirow{2}{*}{\rotatebox{90}{}}  & \multirow{2}{*}{Model}   &  \multirow{2}{*}{JSD ($\downarrow$)}  & \multicolumn{2}{c|}{MMD ($\downarrow$)}        & \multicolumn{2}{c|}{Coverage ($\uparrow$)}    \\ \cline{4-7} 
		&                       &             & CD            & EMD            & CD             & EMD            \\\hline \hline
		        \multirow{5}{*}{\rotatebox{90}{sofa}}         & r-GAN        & 1.866          & 2.037         & 2.247          & 13.00             & 23.00             \\
                                    & l-GAN        & 0.681          & 0.631          & \textbf{1.028} & \textbf{43.00}    & 44.00             \\
                                    & PointFlow    & \textbf{0.244} & 0.585 & 1.313 & 34.00 & 33.00          \\
                                    & Ours         & 0.647          & \textbf{0.547} & 1.089          & 39.00             & \textbf{45.00}    \\\cline{2-7}
                                    & Training Set & 0.185          & 0.467          & 0.904          & 56.00             & 56.00             \\\hline
        \multirow{5}{*}{\rotatebox{90}{bed}}          & r-GAN        & 1.973          & 1.250          & 2.441          & 27.00             & 21.00             \\
                                    & l-GAN        & 0.646          & \textbf{0.539}          & \textbf{0.992}          & 48.00             & 44.00             \\
                                    & PointFlow    & \textbf{0.219} & 0.544 & 1.230 & \textbf{50.00} & 35.00          \\
                                    & Ours         & 0.461          & 0.552          & 1.004 & \textbf{50.00}    & \textbf{50.00}    \\\cline{2-7}
                                    & Training Set & 0.169          & 0.516          & 0.927          & 57.00          & 55.00             \\\hline
        \multirow{5}{*}{\rotatebox{90}{table}}        & r-GAN        & 3.801          & 3.714          & 2.625          & 8.00              & 14.00             \\
                                    & l-GAN        & 4.254          & 1.232           & 2.166          & 14.00             & 9.00              \\
                                    & PointFlow    & 1.044 & 1.630 & 1.535 & 16.00 & 29.00             \\
                                    & Ours         & \textbf{0.869} & \textbf{0.640} & \textbf{1.000} & \textbf{44.00}    & \textbf{37.00}    \\\cline{2-7}
                                    & Training Set & 0.703          & 1.218           & 1.182          & 31.00             & 38.00          \\\hline
        \multirow{5}{*}{\rotatebox{90}{desk}}        & r-GAN        & 3.575          & 2.712          & 3.678          & 22.09          & 22.09          \\
                                    & l-GAN        & 2.233          & \textbf{1.139} & 2.345          & 38.37          & 25.58          \\
                                    & PointFlow    & \textbf{0.327} & 1.254 & \textbf{1.548} & 38.37 & 46.51             \\
                                    & Ours         & 0.454          & 1.223          & 1.567          & \textbf{56.98} & \textbf{52.33} \\\cline{2-7}
                                    & Training Set & 0.329          & 1.055          & 1.332          & 53.48          & 50.00             \\\hline
        \multirow{5}{*}{\rotatebox{90}{dresser}}      & r-GAN        & 1.726          & 1.299          & 1.675           & 36.05          & 30.23          \\
                                    & l-GAN        & 0.648          & 0.642          & 1.010          & 45.35          & 43.02          \\
                                    & PointFlow    & \textbf{0.270} & 0.715 & 1.349 & 46.51 & 37.21          \\
                                    & Ours         & 0.457          & \textbf{0.485}  & \textbf{0.988} & \textbf{53.49} & \textbf{52.33} \\\cline{2-7}
                                    & Training Set & 0.215          & 0.551          & 0.882           & 56.98           & 54.65         \\\hline
	\end{tabular}
	\caption{Comparison of quality of point cloud synthesis on the ModelNet10. $\downarrow$: the lower the better, $\uparrow$: the higher the better. MMD-CD scores are multiplied by 100; MMD-EMD scores and JSDs are multiplied by 10.}
	\label{tab:generation}  
\end{table*}    

\section{Experiments}

We conduct experiments to test the proposed GPointNet model for point cloud modeling on a variety of tasks below. The code and more results can be found at: \url{http://www.stat.ucla.edu/~jxie/GPointNet}.

\begin{table}[b]
	\small
	\centering
	\setlength{\tabcolsep}{1mm}{
		\begin{tabular}{|l|c|cc|c|cc|}
			\hline 
			Model                     & Category  & CD              & EMD & Category  & CD              & EMD               \\
			\hline \hline
			Ours & \multirow{2}{*}{night stand}  & \textbf{0.378} & \textbf{0.685} & \multirow{2}{*}{sofa}  & 0.427 &\textbf{0.703}       \\
			                              PointFlow &  & 0.464          & 0.990   &  & \textbf{0.389} & 0.888       \\\hline
			Ours & \multirow{2}{*}{toilet}      & \textbf{0.396} & \textbf{0.708} & \multirow{2}{*}{bed}        & \textbf{0.361} & \textbf{0.670} \\
			PointFlow &  							& 0.456          & 0.992  &  & 0.372          & 0.914           \\\hline
			Ours & \multirow{2}{*}{monitor}      & \textbf{0.371} & \textbf{0.705} & \multirow{2}{*}{table}          & \textbf{0.318} & \textbf{0.621}\\
			PointFlow &  & 0.441          & 0.957  &  & 0.581          & 1.008        \\\hline
			Ours & \multirow{2}{*}{chair}         & \textbf{0.337} & \textbf{0.719} & \multirow{2}{*}{desk}       & \textbf{0.391} & \textbf{0.697} \\
			PointFlow &  & 0.510          & 1.028 &  & 0.500          & 1.063         \\\hline
			Ours & \multirow{2}{*}{bathtub}      & 0.321  & \textbf{0.612} & \multirow{2}{*}{dresser}        & \textbf{0.329}  & \textbf{0.645} \\
			PointFlow &  & \textbf{0.289}          & 0.825 &  & 0.415          & 0.942         \\\hline
		\end{tabular}
	}
	\caption{Comparison of performance in reconstruction on the ModelNet10. CD scores are multiplied by 100 and EMD scores are multiplied by 10. The lower the better.}
	\label{tab:rec}
\end{table}

\begin{table*}[!h]
	\begin{minipage}[]{0.65\textwidth}
		\centering	
        \setlength{\tabcolsep}{1pt}
        \renewcommand{\arraystretch}{0.5}
        \begin{tabular}{ccccccccc} 
        \hspace{0.5mm}\rotatebox{90}{\hspace{3mm}{\footnotesize Input}} &
		\includegraphics[trim=30 30 30 30,clip,width=0.11\linewidth]{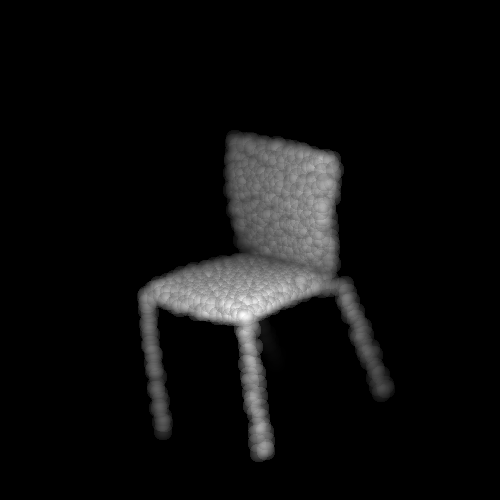}  &
		\includegraphics[trim=30 30 30 30,clip,width=0.11\linewidth]{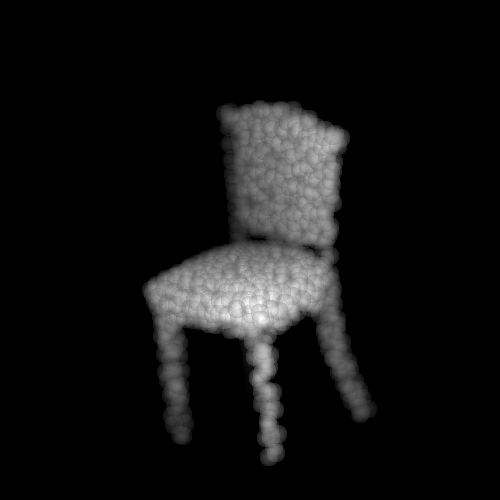}  &
	 	\includegraphics[trim=50 30 50 70,clip,width=0.11\linewidth]{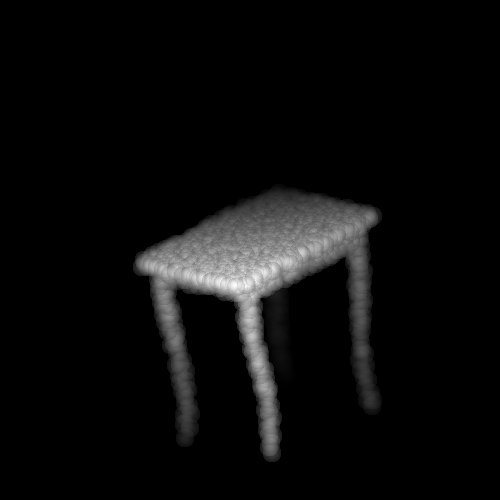} &
	 	\includegraphics[trim=50 30 50 70,clip,width=0.11\linewidth]{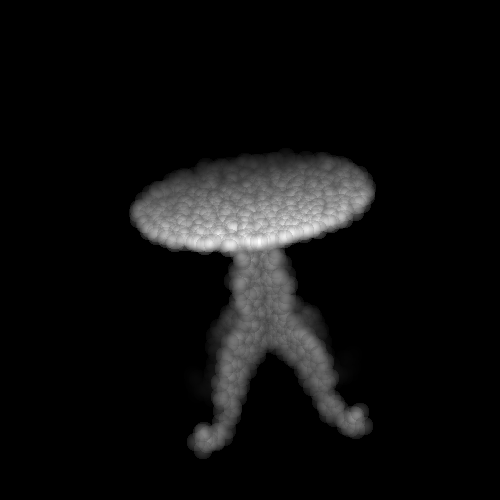} &
		\includegraphics[trim=30 30 30 30,clip,width=0.11\linewidth]{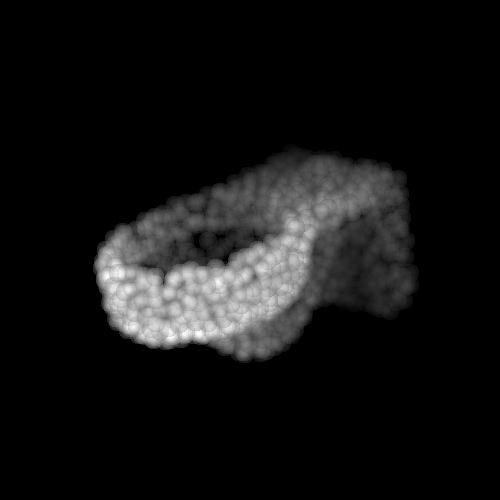}  &
		\includegraphics[trim=30 30 30 30,clip,width=0.11\linewidth]{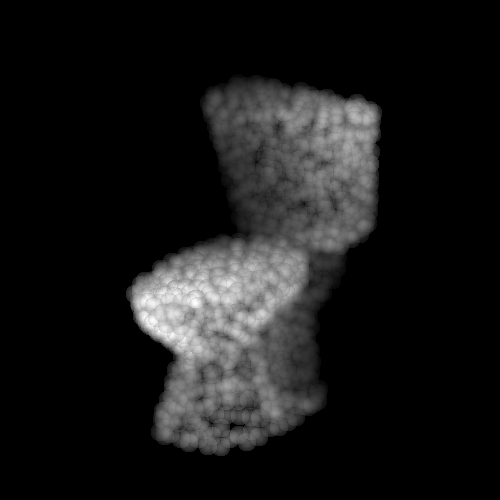}  &
		\includegraphics[trim=30 30 30 30,clip,width=0.11\linewidth]{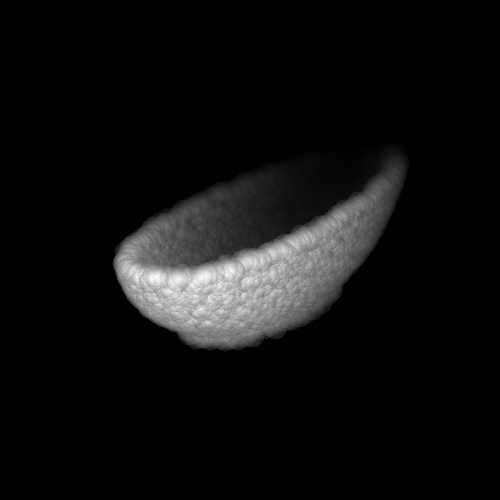} &
		\includegraphics[trim=30 30 30 30,clip,width=0.11\linewidth]{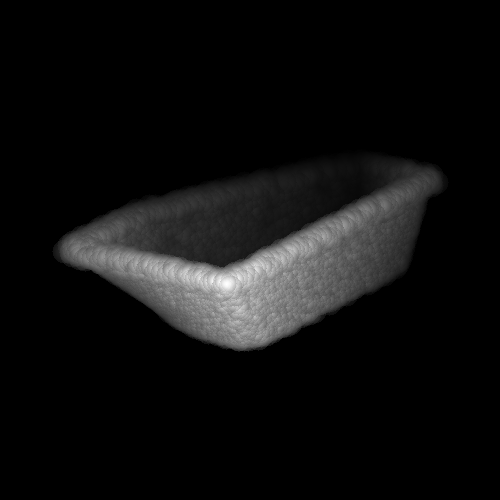} 
		\\ 
		\hspace{0.5mm}\rotatebox{90}{\hspace{4mm}{\footnotesize Ours}} &
  	\includegraphics[trim=30 30 30 30,clip,width=0.11\linewidth]{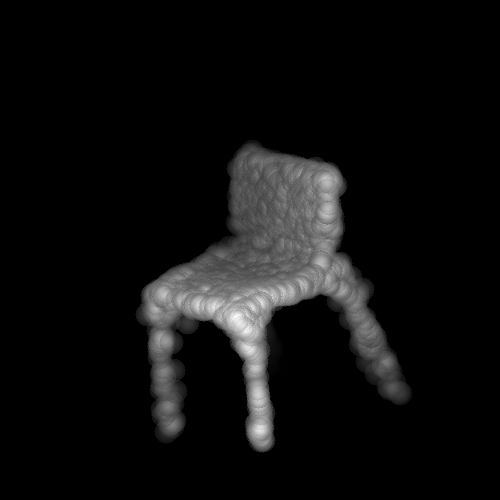} &
		\includegraphics[trim=30 30 30 30,clip,width=0.11\linewidth]{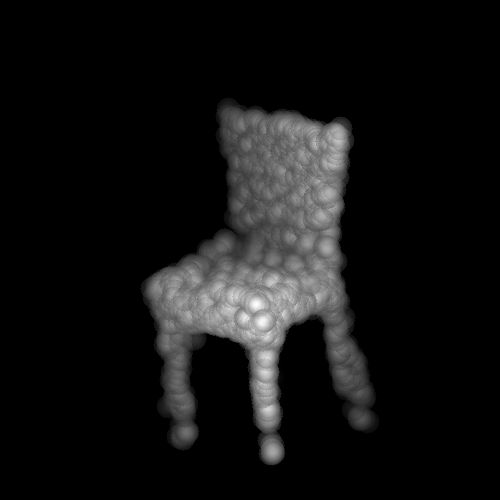} &
	 	\includegraphics[trim=50 30 50 70,clip,width=0.11\linewidth]{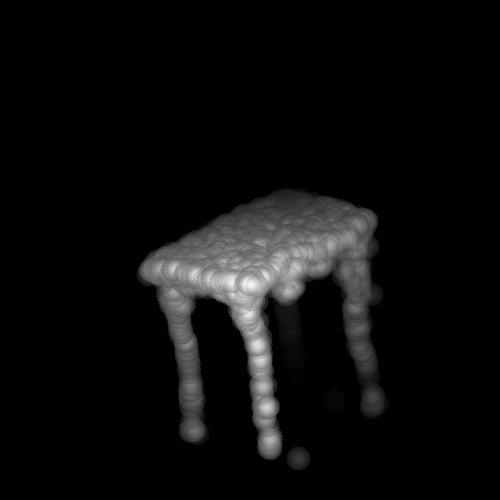} &
	 	\includegraphics[trim=50 30 50 70,clip,width=0.11\linewidth]{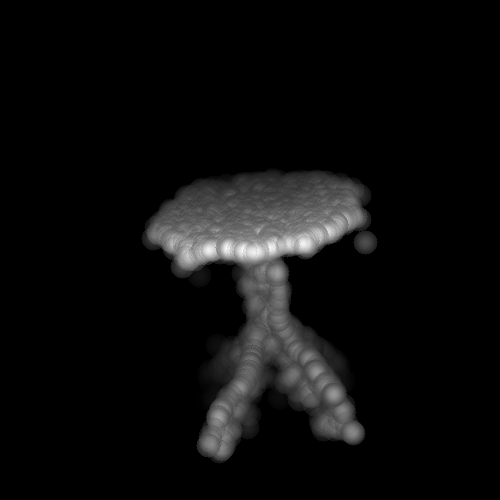} &
		\includegraphics[trim=30 30 30 30,clip,width=0.11\linewidth]{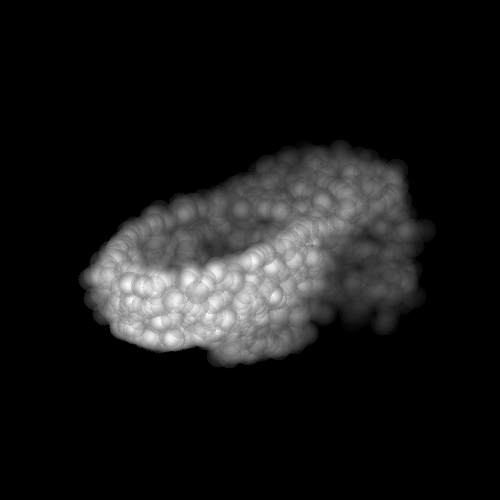} &
		\includegraphics[trim=30 30 30 30,clip,width=0.11\linewidth]{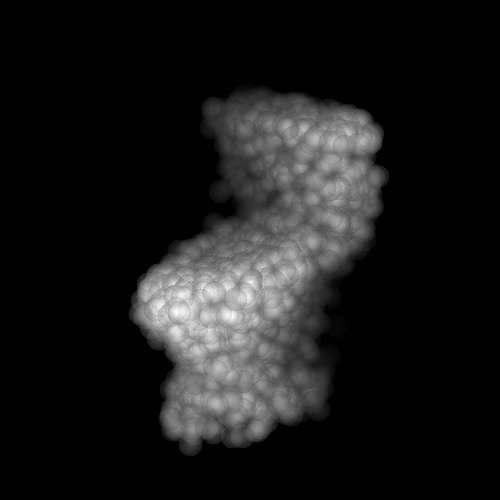} &
		\includegraphics[trim=30 30 30 30,clip,width=0.11\linewidth]{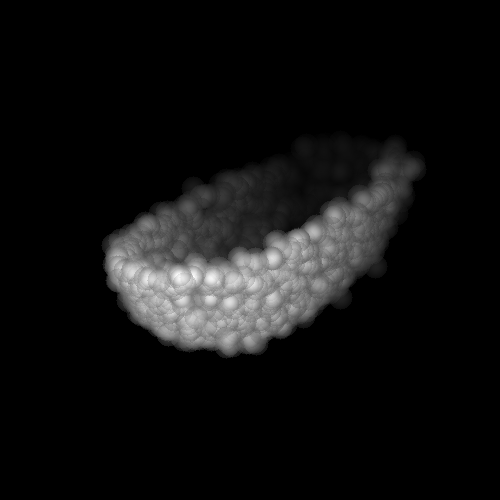} &
		\includegraphics[trim=30 30 30 30,clip,width=0.11\linewidth]{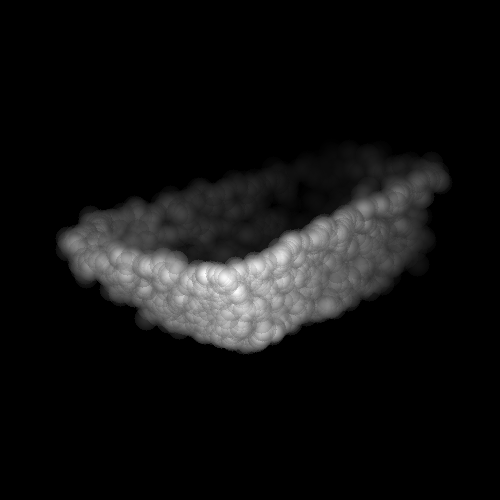}  
		\\ 
		\hspace{0.5mm}\rotatebox{90}{\hspace{0.3mm}{\footnotesize PointFlow}} &
		\includegraphics[trim=30 30 30 30,clip,width=0.11\linewidth]{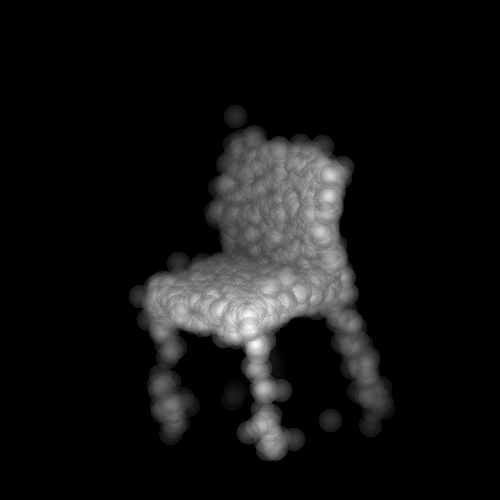} &
		\includegraphics[trim=30 30 30 30,clip,width=0.11\linewidth]{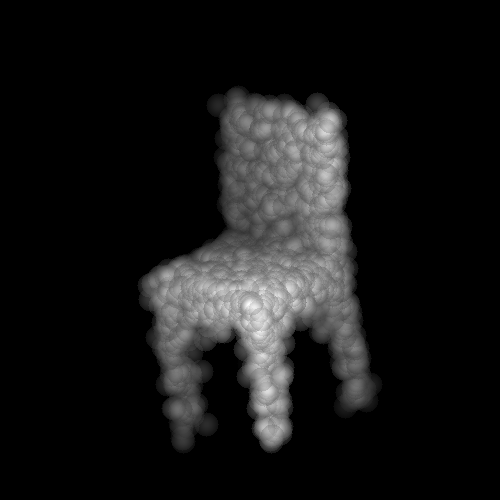} &
	 	\includegraphics[trim=50 30 50 70,clip,width=0.11\linewidth]{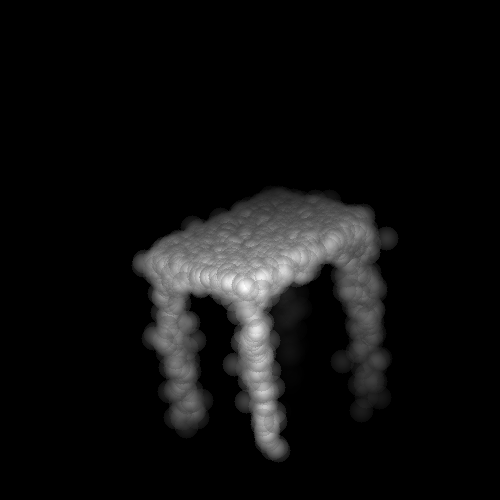} &
	 	\includegraphics[trim=50 30 50 70,clip,width=0.11\linewidth]{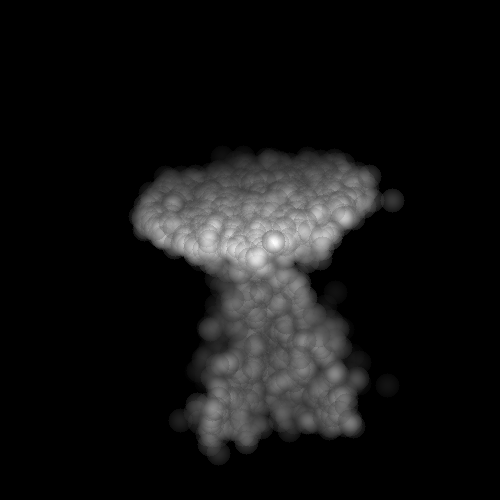} &
		\includegraphics[trim=30 30 30 30,clip,width=0.11\linewidth]{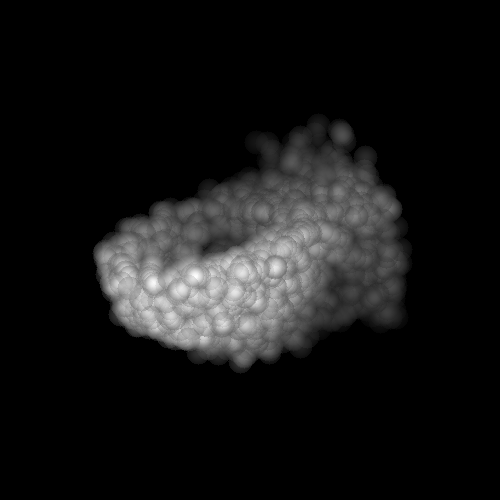} &
		\includegraphics[trim=30 30 30 30,clip,width=0.11\linewidth]{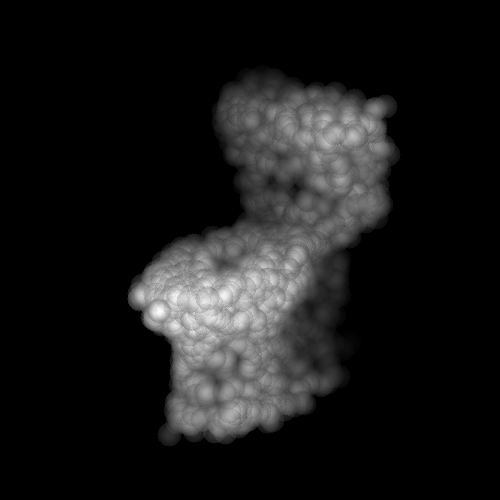} &
		\includegraphics[trim=30 30 30 30,clip,width=0.11\linewidth]{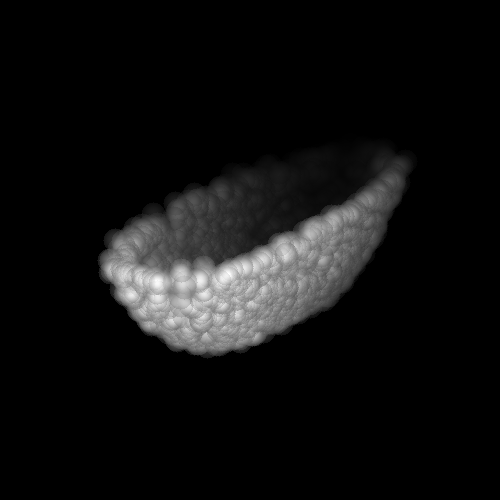} &
		\includegraphics[trim=30 30 30 30,clip,width=0.11\linewidth]{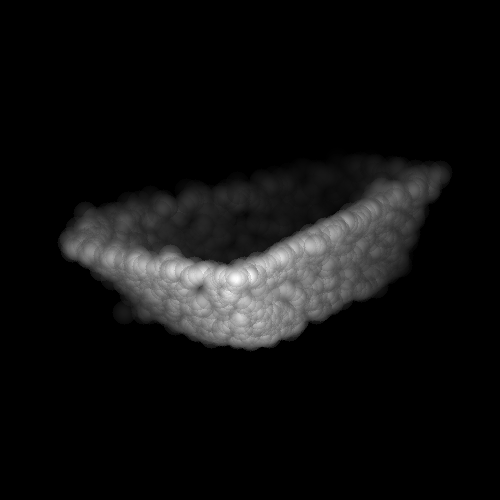} 
            \end{tabular}   
		\captionof{figure}{Point cloud reconstruction.~A short-run MCMC as a generator is learned from chair, table, toilet and bathtub,  respectively. The learned generator is applied to reconstruction by inferring the latent $Z$ to minimize the reconstruction error.} 
		\label{fig:rec}
         \vspace{3mm}
    	\centering	
        \setlength{\tabcolsep}{1pt}
        \renewcommand{\arraystretch}{0.5}
        \begin{tabular}{ccccccccc} 
                
            \hspace{0.5mm}\rotatebox{90}{\hspace{3mm}{\footnotesize Chair}} &
            \includegraphics[trim=30 30 30 30,clip,width=0.11\linewidth]{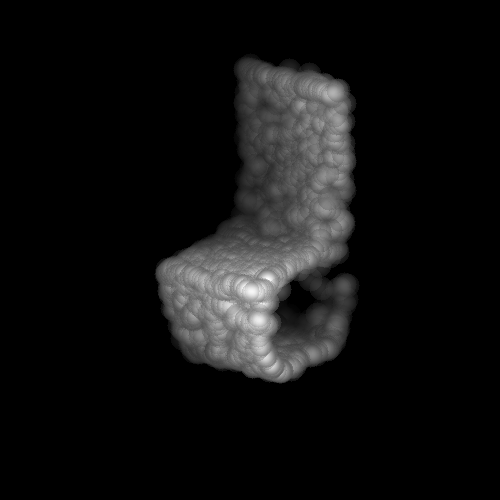} &
            \includegraphics[trim=30 30 30 30,clip,width=0.11\linewidth]{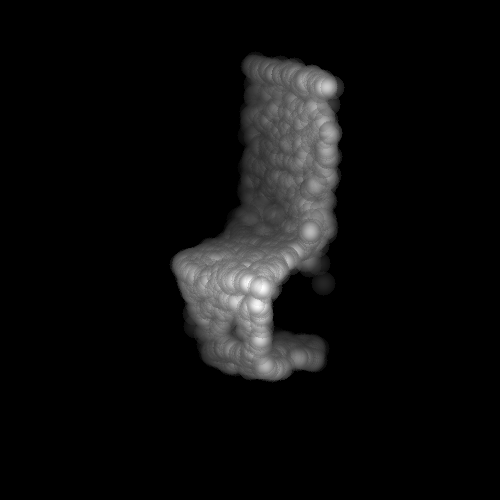} &
            \includegraphics[trim=30 30 30 30,clip,width=0.11\linewidth]{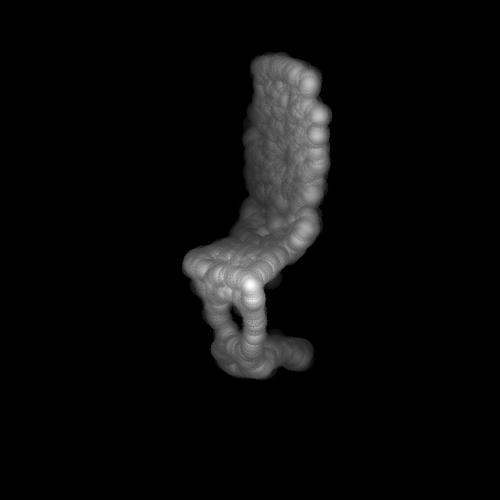} &
            \includegraphics[trim=30 30 30 30,clip,width=0.11\linewidth]{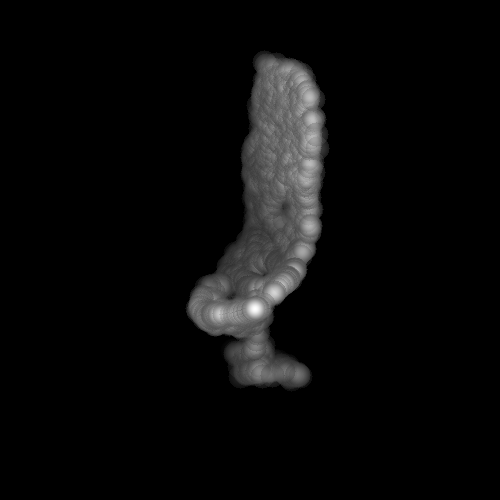} &
            \includegraphics[trim=30 30 30 30,clip,width=0.11\linewidth]{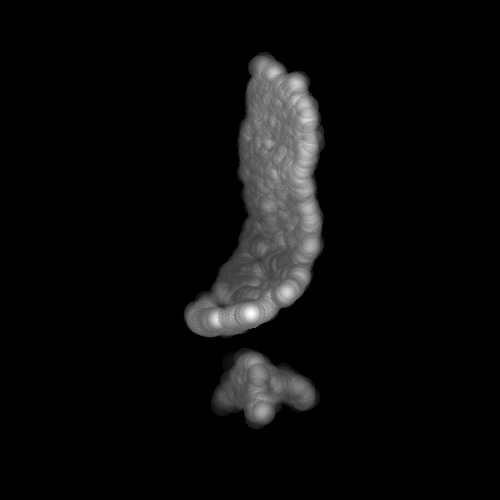} &
            \includegraphics[trim=30 30 30 30,clip,width=0.11\linewidth]{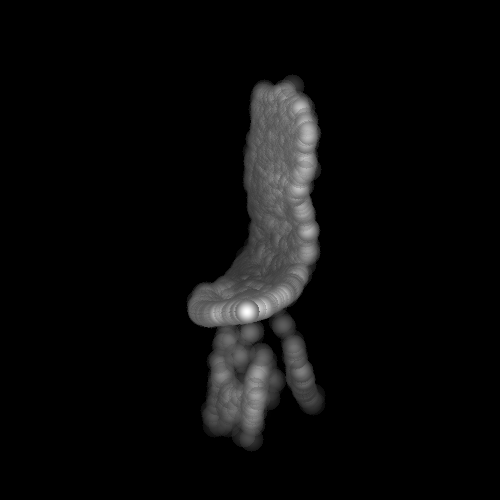} &
            \includegraphics[trim=30 30 30 30,clip,width=0.11\linewidth]{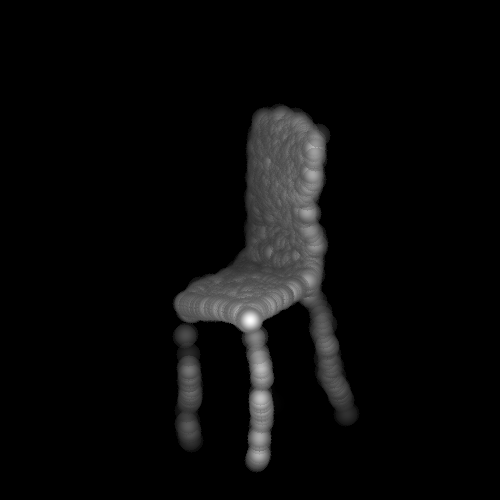} &
            \includegraphics[trim=30 30 30 30,clip,width=0.11\linewidth]{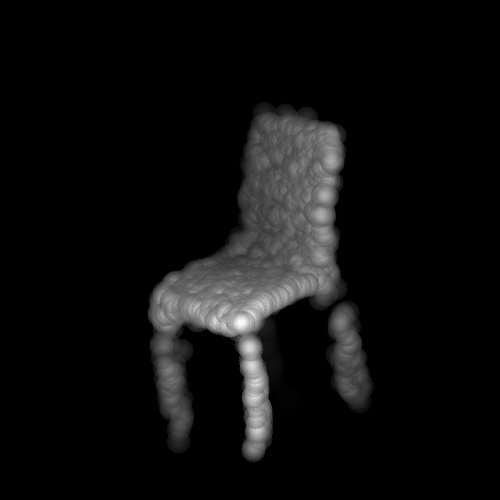}\\ 
            \hspace{0.5mm}\rotatebox{90}{\hspace{3mm}{\footnotesize Toilet}} &
            \includegraphics[trim=60 60 60 60,clip,width=0.11\linewidth]{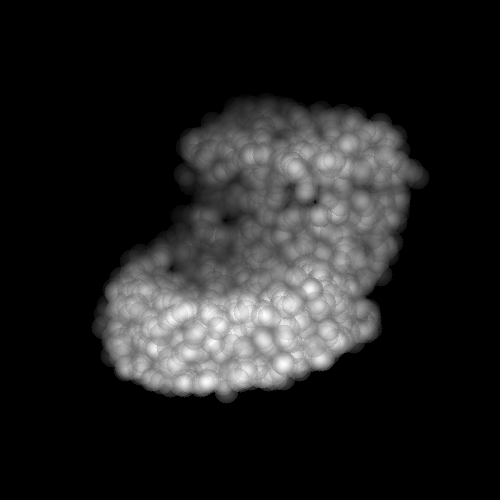} &
            \includegraphics[trim=60 60 60 60,clip,width=0.11\linewidth]{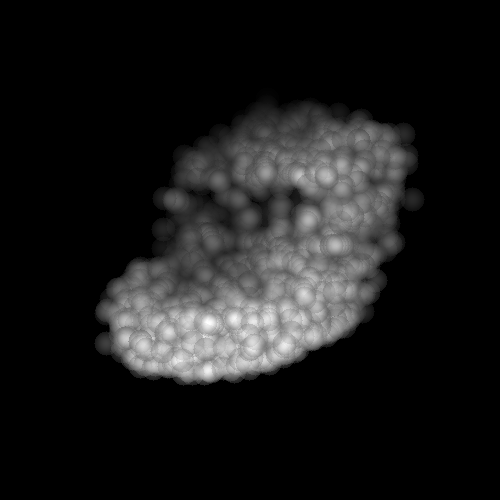} &
            \includegraphics[trim=60 60 60 60,clip,width=0.11\linewidth]{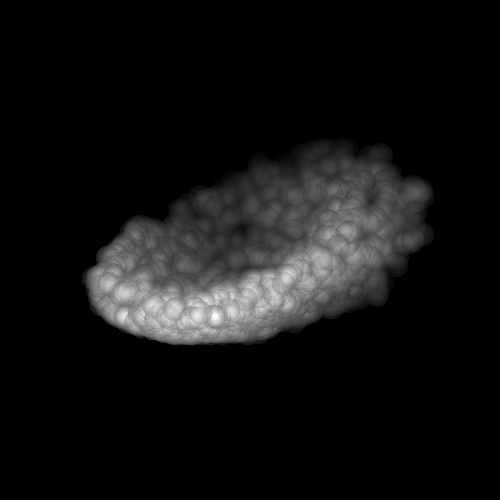} &
            \includegraphics[trim=60 60 60 60,clip,width=0.11\linewidth]{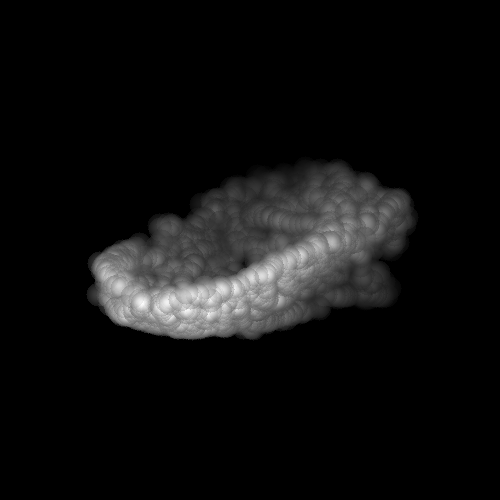} &
            \includegraphics[trim=60 60 60 60,clip,width=0.11\linewidth]{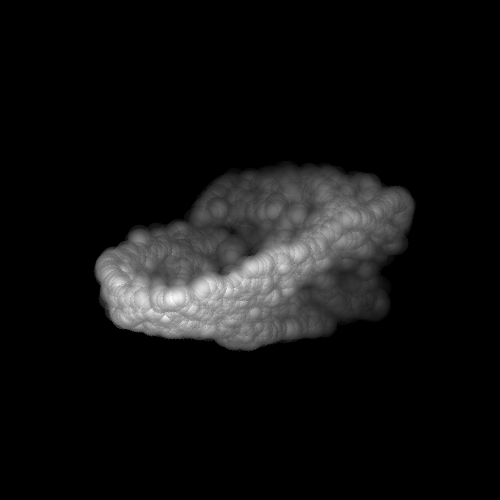} &
            \includegraphics[trim=60 60 60 60,clip,width=0.11\linewidth]{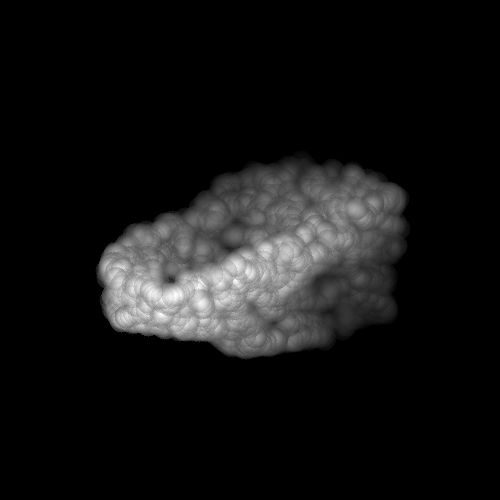} &
            \includegraphics[trim=60 60 60 60,clip,width=0.11\linewidth]{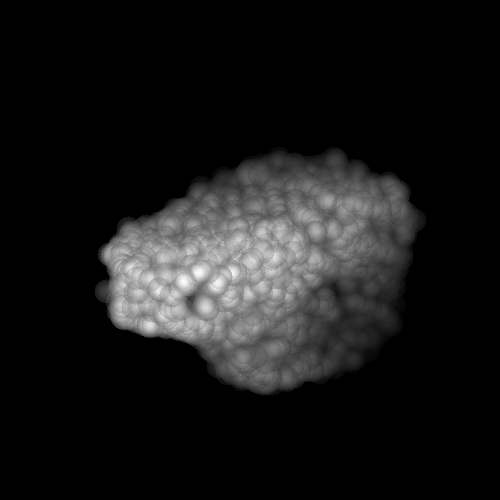} &
            \includegraphics[trim=60 60 60 60,clip,width=0.11\linewidth]{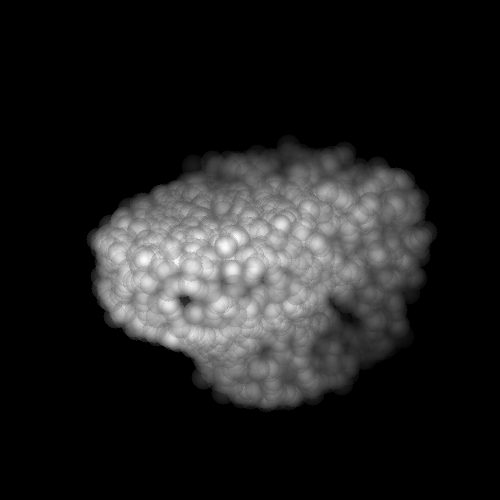}   
            \end{tabular}   
	\caption{Point cloud interpolation between the generated examples at two ends. The transition in each row displays the sequence of $M_{\theta}(Z_{\rho})$ with the linear interpolated latent variable $Z_{\rho}=\rho Z_1 + (1-\rho) Z_2$, where $\rho \in [0,1]$. The left and right point clouds are $M_{\theta}(Z_1)$ and $M_{\theta}(Z_2)$, respectively. 
	}
	\label{fig:interpolation}
	\end{minipage}
	\hfill 
	\begin{minipage}[]{0.31\textwidth}
		\small
		\centering
		\setlength{\tabcolsep}{3.2mm}{
			\begin{tabular}{|l|cc|}
				\hline 
				Method & Full & Generation \\ \hline \hline
				r-GAN  & 7.22 & 6.91 \\
				l-GAN & 1.97 & 1.71 \\
				PointFlow & 1.61 & 1.06 \\\hline
				Ours &  \multicolumn{2}{c|}{1.39} \\ \hline
				
			\end{tabular}
		}
		\caption{A comparison of model sizes. Our method has only one network for both learning and generation.(Million)}
		\label{tab:parameter} 
         \vspace{3mm}
		\begin{tabular}{|l|c|}
			\hline
			Method    & Accuracy \\ \hline \hline
			SPH \cite{kazhdan2003rotation} & 79.8\% \\
			LFD \cite{chen2003visual} & 79.9\% \\
			PANORAMA-NN \cite{SfikasTP17} & 91.1\%\\
			VConv-DAE \cite{sharma2016vconv} & 80.5\% \\
			3D-GAN \cite{wu2016learning} & 91.0\% \\
			3D-WINN \cite{huang20193d} & 91.9\% \\
			3D-DescriptorNet \cite{xie2018learning} & 92.4\% \\
			Primitive GAN \cite{khan2019unsupervised} & 92.2\%\\
			FoldingNet \cite{yang2018foldingnet} & 94.4\% \\
			l-GAN \cite{achlioptas2018learning} & 95.4\% \\
			PointFlow \cite{yang2019pointflow} & 93.7\% \\ \hline
			Ours & 93.7\% \\ \hline
		\end{tabular}
		\caption{A comparison of accuracy of 3D object classification on the 10-category ModelNet10 dataset.}
		\label{tab:classification}
	\end{minipage}
\end{table*}

\subsection{Synthesis}
We evaluate our model for 3D point cloud synthesis on the ModelNet10, a 10-category subset of ModelNet \cite{wu20153d} which is commonly used as a benchmark for 3D object analysis. We first create a point cloud dataset by sampling points uniformly from the mesh surface of each object in the ModelNet10 dataset, and then scale them into a range of [-1,1].
We train one single model for each category of point clouds. The number of training examples in each category ranges from 100 to 900. Each point cloud contains 2,048 points.

The network structure of the scoring function $f_{\theta}(X)$ is visualized in Figure \ref{fig:structure}. It first encodes each 3-dimensional point coordinate in Euclidean space to a 1,024-dimensional point feature by an MLP, then uses an average pooling layer to aggregate information from all the points to a single 1,024-dimensional global point cloud feature, and maps it to the score by another MLP. The scoring function is input-permutation-invariant because the MLP for point encoding is shared by all unordered points and also the output of the symmetric function, which is an average pooling layer followed by an MLP, is not affected by the point feeding order. 

We use Adam \cite{KingmaB14adam} for optimization with an initial learning rate 0.005, $\beta_1 = 0.9$ and $\beta_2 = 0.999$. We decay the learning rate by 0.985 for every 50 iterations. The minibatch size is 128. The number of paralleled MCMC chains is 128. We run $K$ = 64 Langevin steps, with the step size $\delta=0.005$. To avoid exploding gradients in MCMC, we clip the gradient values to a range [-1,1] at each Langevin step. We run 2,000 iterations for training. 
To further improve training, we inject additive Gaussian noises with stand deviation 0.01 to the observed examples at each iteration.

To quantitatively evaluate the performance of generative models of point clouds, we adopt three metrics that are also used in \cite{achlioptas2018learning,yang2019pointflow}, i.e.,
Jensen-Shannon Divergence (JSD), Coverage (COV) and Minimum Matching Distance (MMD). When evaluating COV and MMD, two point clouds are measured by either Chamfer distance (CD) or earth mover's distance (EMD). 
We compare our model with some baseline generative models for point clouds, including PointFlow \cite{yang2019pointflow}, l-GAN, and r-GAN, in Table \ref{tab:generation}. We report the performance of the baselines using their official codes. 
Figure \ref{fig:syn} displays some examples of point clouds generated by our model for categories chair, toilet, table, and bathtub.

\subsection{Reconstruction}

We demonstrate the reconstruction ability of the GPointNet~model for 3D point clouds. We learn our model with a short-run MCMC as a generator. Given a testing point cloud object, we reconstruct it with the learned generator by minimizing the reconstruction error as we discussed in Section \ref{sec:generator}. Figure \ref{fig:rec} displays some examples of reconstructing unobserved examples. The first row displays the original point clouds to reconstruct, the second row shows the corresponding reconstructed point clouds obtained by the learned model, and the third row shows the results obtained by a baseline, PointFlow \cite{yang2019pointflow}, which is a VAE-based framework. For VAE, the reconstruction can be easily achieved by first inferring the latent variables of the input example and then mapping the inferred latent variables back to the point cloud space via the generator. Table \ref{tab:rec} shows a quantitative comparison of our method with PointFlow for point cloud reconstruction. CD and EMD metrics are adopted to measure the quality of the reconstruction. On the whole, our method outperforms the baseline.

As to model complexity, we also compare the numbers of parameters of different models in Table \ref{tab:parameter}. Due to the usage of extra networks in learning, models based on GAN and VAE have different sizes of parameters in training and generation stages. Our model does not use an auxiliary network, thus it has less parameters. 

\subsection{Interpolation}
We demonstrate the interpolation ability of our model. We learn the model with short-run MCMC. We first sample two noise point clouds $Z_1$ and $Z_2$ from Gaussian distribution as two samples from the latent space. Then we perform linear interpolation in the latent space $Z_{\rho}= (1- \rho)\cdot Z_1 + \rho \cdot Z_2$, with $\rho$ discretized into 8 values within $[0,1]$. We generate point clouds by $X_{\rho} = M_{\theta}(Z_{\rho})$. Figure \ref{fig:interpolation} shows two results of interpolation between $Z_1$ and $Z_2$ by showing the sequences of generated point clouds $\{X_{\rho}\}$. Smooth transition and physically plausible intermediate generated examples suggest that the generator learns a smooth latent space for point~cloud~embedding.  

\subsection{Representation learning for classification}

The learned point encoder $h(x)$ in the scoring function $f_{\theta}(X)$ can be useful for point cloud feature extraction, and the features can be applied to supervised learning. We evaluate $h$ by performing a classification experiment on the ModelNet10 dataset. We first train a single GPointNet on the training examples from all categories in an unsupervised manner. The network $f_{\theta}(X)$ is the same as the one used in the previous sections, except that we add one layer with 2,048 channels before the average pooling and one layer with 1,024 channels after the average pooling. We replace the average pooling layer by a max-pooling layer in the learned scoring function and use the output of the max-pooling as point cloud features. Such a point cloud feature extractor is also permutation-invariant. We train an SVM \cite{svmbook} classifier from labeled data based on the extracted features for classification. We evaluate the classification accuracy of the SVM on the testing data using the one-versus-all rule.
Table \ref{tab:classification} reports 11 published results on this dataset obtained by other baselines. Our method is on a par with other methods in terms of classification accuracy on this dataset.

\begin{figure}[h]
	\centering	
	\includegraphics[trim=0 0 45 40,clip,width=0.324\linewidth]{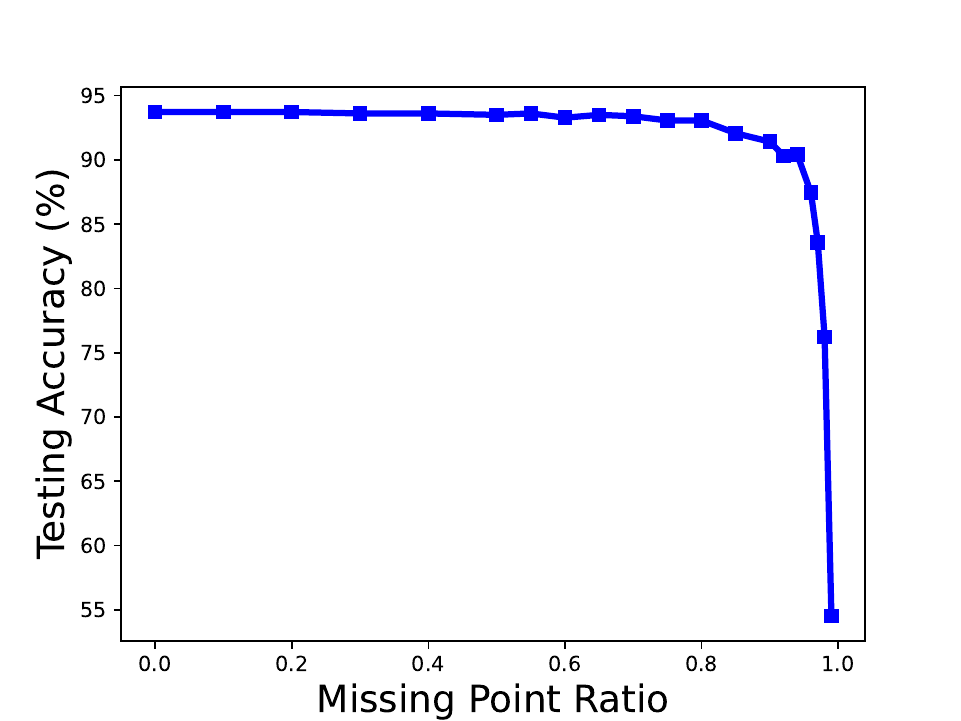}
	\includegraphics[trim=0 0 45 40,clip,width=0.324\linewidth]{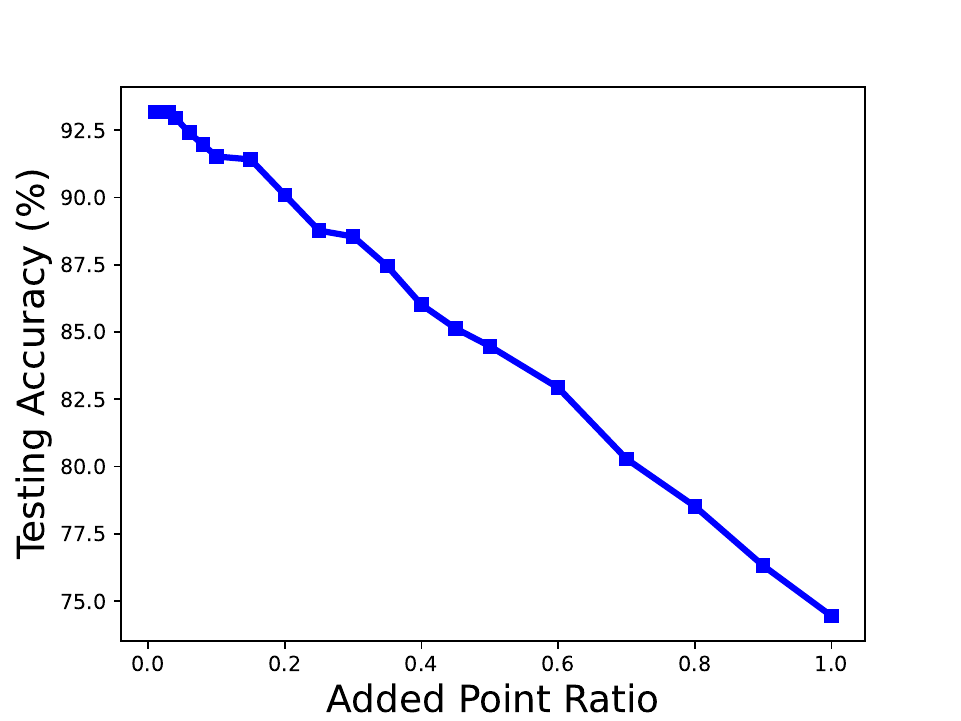}
	\includegraphics[trim=15 0 30 40,clip,width=0.324\linewidth]{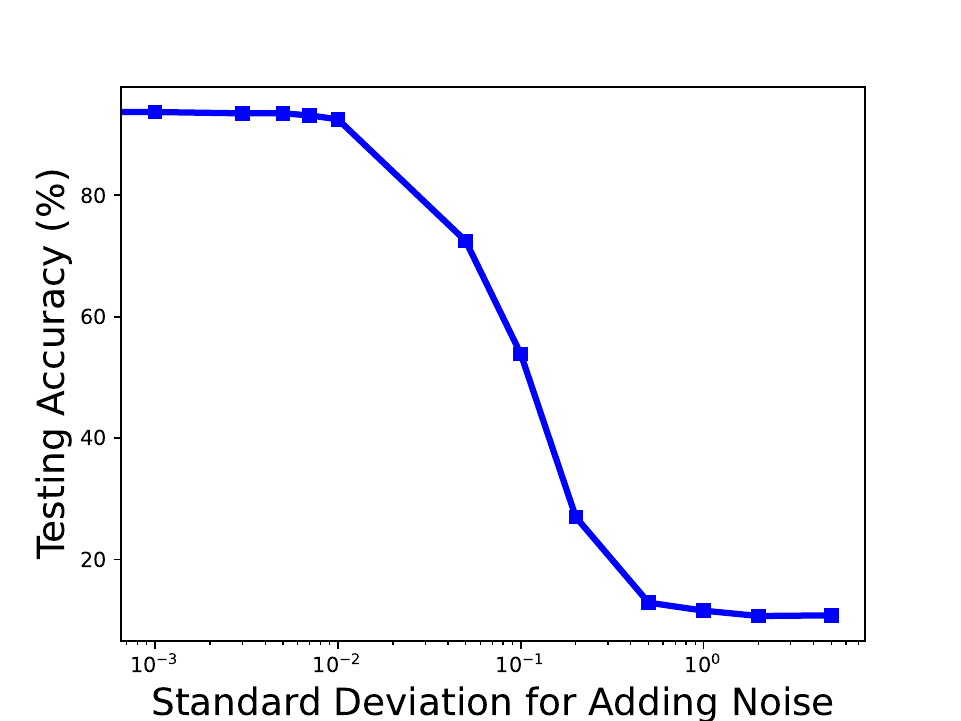}
	\caption{Robustness Test. The model is tested on ModelNet10 test set with three types of point corruptions. Classification accuracies are reported across different levels of corruptions. Left: missing points. Middle: added points. Right: point perturbation.}
	\label{fig:robustness}
\end{figure}

We conduct experiments to test the robustness of the classifier. We consider the following three types of data corruptions: (1) Type 1: missing points, where we randomly delete points from each point cloud. (2) Type 2: added points, where we add extra points that are uniformly distributed in the cube $[-1, 1]^3$ into each point cloud. (3) Type 3: point perturbation, where we perturb each point of each point cloud by adding a Gaussian noise. We report classification accuracy of the classifier on the corrupted version of ModelNet10 test set. Figure \ref{fig:robustness} shows the results. The classification performance decreases as the corruption level (e.g., missing point ratio, added point ratio, and standard deviation of point perturbation) increases. In the case of missing points, even though $94\%$ points are deleted in each testing example, the classifier can still perform with an accuracy $90.20\%$. In the extreme case where we only keep 20 points ($1\%$) in each point cloud, the accuracy becomes $53.19\%$.

\subsection{Visualization of point encoding function}
\label{sec:point_fun}
\begin{figure}[h]
	\centering	
\includegraphics[trim=18 18 18 18,clip,width=0.156\linewidth]{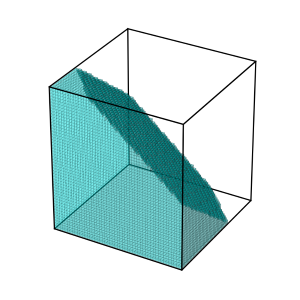}
\includegraphics[trim=18 18 18 18,clip,width=0.156\linewidth]{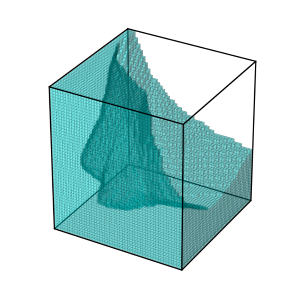}
\includegraphics[trim=18 18 18 18,clip,width=0.156\linewidth]{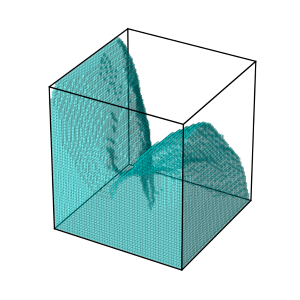}
\includegraphics[trim=18 18 18 18,clip,width=0.156\linewidth]{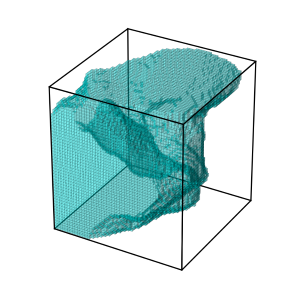}
\includegraphics[trim=18 18 18 18,clip,width=0.156\linewidth]{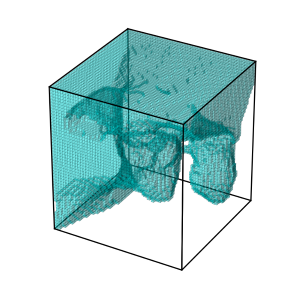}
\includegraphics[trim=18 18 18 18,clip,width=0.156\linewidth]{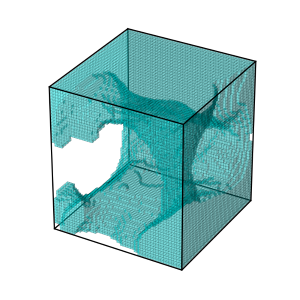}
\includegraphics[trim=18 18 18 18,clip,width=0.156\linewidth]{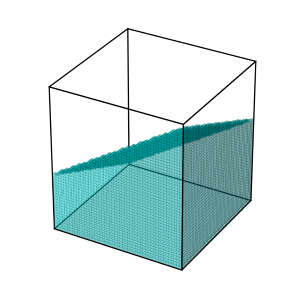}
\includegraphics[trim=18 18 18 18,clip,width=0.156\linewidth]{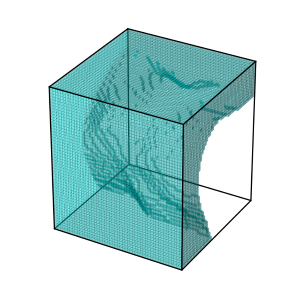}
\includegraphics[trim=18 18 18 18,clip,width=0.156\linewidth]{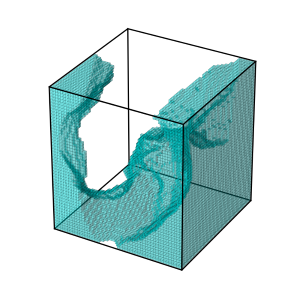}
\includegraphics[trim=18 18 18 18,clip,width=0.156\linewidth]{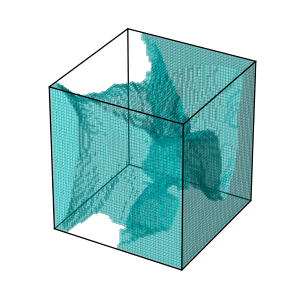}
\includegraphics[trim=18 18 18 18,clip,width=0.156\linewidth]{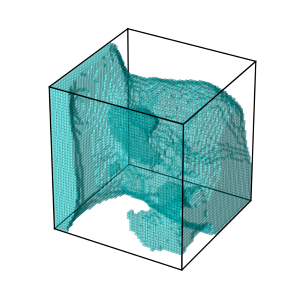}
\includegraphics[trim=18 18 18 18,clip,width=0.156\linewidth]{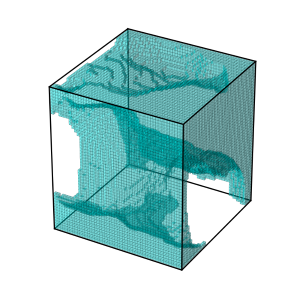}
\includegraphics[trim=18 18 18 18,clip,width=0.156\linewidth]{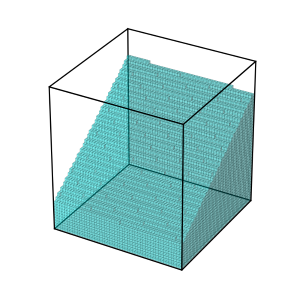}
\includegraphics[trim=18 18 18 18,clip,width=0.156\linewidth]{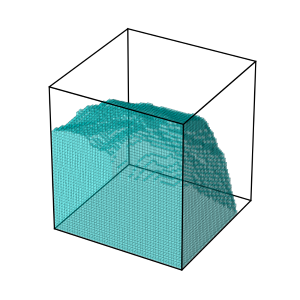}
\includegraphics[trim=18 18 18 18,clip,width=0.156\linewidth]{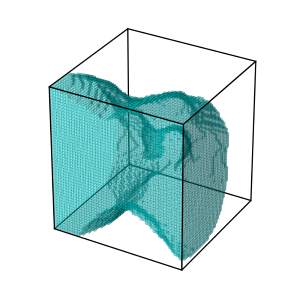}
\includegraphics[trim=18 18 18 18,clip,width=0.156\linewidth]{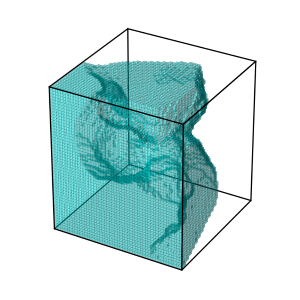}
\includegraphics[trim=18 18 18 18,clip,width=0.156\linewidth]{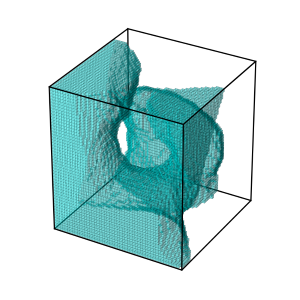}
\includegraphics[trim=18 18 18 18,clip,width=0.156\linewidth]{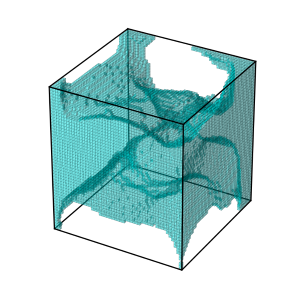}
\includegraphics[trim=18 18 18 18,clip,width=0.156\linewidth]{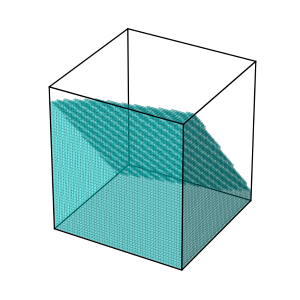}
\includegraphics[trim=18 18 18 18,clip,width=0.156\linewidth]{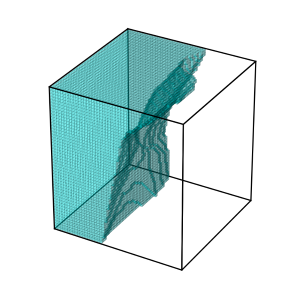}
\includegraphics[trim=18 18 18 18,clip,width=0.156\linewidth]{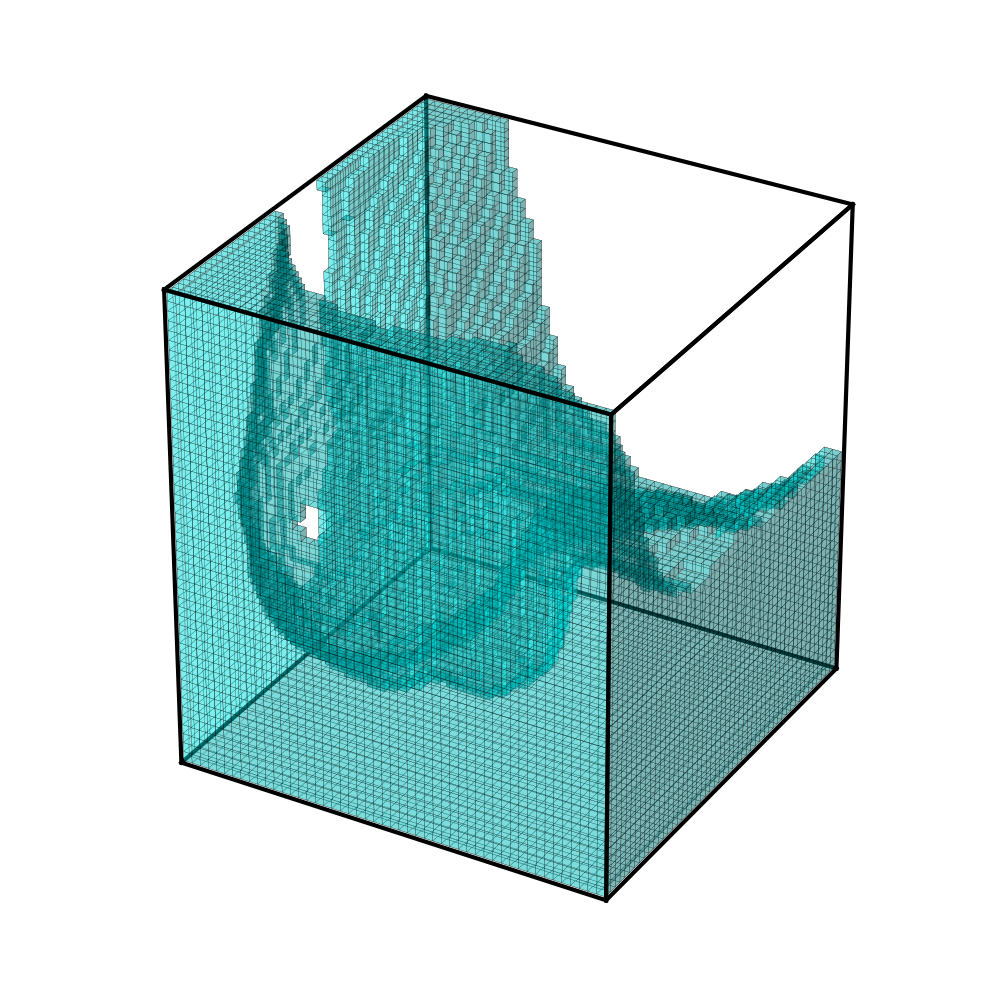}
\includegraphics[trim=18 18 18 18,clip,width=0.156\linewidth]{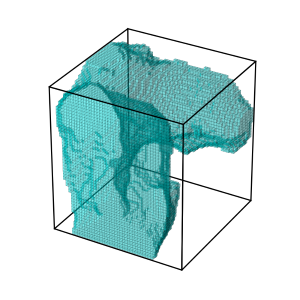}
\includegraphics[trim=18 18 18 18,clip,width=0.156\linewidth]{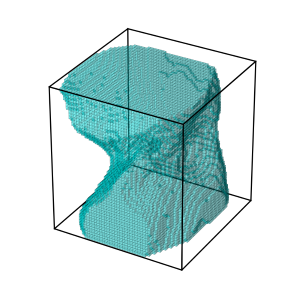}
\includegraphics[trim=18 18 18 18,clip,width=0.156\linewidth]{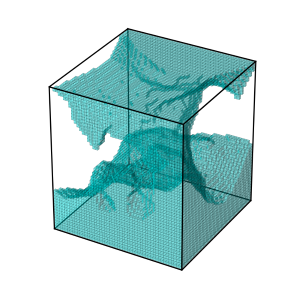}
	\\ layer 1 \hspace{2mm} layer 2 \hspace{2mm} layer 3 \hspace{2mm} layer 4 \hspace{2mm} layer 5 \hspace{2mm} layer 6
	\caption{Visualization of point encoding functions. The point encoding function is implemented by a MLP. Each filter at different layers of the MLP is visualized by showing the points that have positive filter responses. Four filters randomly selected at each layer are visualized. }
	\label{fig:pint_function_vis}
\end{figure}

The scoring function learns a coordinate encoding of each point and then aggregates all individual point codes into a score for the point set.  The coordinate encoding function is implemented by an MLP, learning to encode each 3-dimensional point to a 2,048-dimensional vector in the model that we use for classification. To better understand what each encoding function learns, we visualize each filter at different layers of the MLP by showing the points in the point cloud domain that give positive filter responses. In Figure \ref{fig:pint_function_vis}, we randomly visualize 4 filters at each layer. The results suggest that different filters at different layers learn to detect points in different shapes of regions. Filters at a higher layer usually detect points in regions with more complicated shapes than those at a lower layer.

\section{Conclusion}
This paper studies the deep energy-based modeling and learning of unordered 3D point clouds. We propose a probability density of 3D point clouds, which is unordered point sets, in the form of the energy-based model where the energy function is parameterized by an input-permutation-invariant deep neural network. The model can be trained via MCMC-based maximum likelihood learning, without the need of recruiting any other assisting network. The learning process follows ``analysis by synthesis'' scheme.  Experiments show that the model can be useful for 3D generation, reconstruction, interpretation, and classification.  

\section*{Acknowledgment}
The work is supported by NSF DMS-2015577,  DARPA XAI project N66001-17-2-4029, ARO project W911NF1810296, ONR MURI project N00014-16-1-2007, and XSEDE grant ASC180018. We thank Erik Nijkamp for insightful discussions about short-run MCMC for EBM and neural tangent kernel. 

{\small
\bibliographystyle{ieee_fullname}
\bibliography{egbib}
}

\end{document}